%% file: paper.tex
\documentclass[]{unifiedreward}
\usepackage{fix-cm}
\usepackage{helvet}           

\usepackage{nicefrac}         
\usepackage{siunitx}         

\usepackage{longtable}
\usepackage{booktabs}

\usepackage{tabularx}         
\usepackage{array}           
\usepackage{makecell}         
\usepackage{threeparttable}  
\usepackage{xspace}

\usepackage{wrapfig}          
\usepackage{float}            
\usepackage[export]{adjustbox}

\usepackage{tikz}             
\usepackage{pgfplots}         
\usepackage{pgf-pie}

\usepackage{listings}        
\usepackage{ragged2e}        
\usepackage{comment}

\usepackage{pifont}           
\usepackage{fontawesome}

\usepackage{enumitem}        
\usepackage{titletoc}         
\usepackage{minitoc}          
\usepackage[toc,page,header]{appendix}

\usepackage{url}              
\usepackage[hang,flushmargin]{footmisc}  
\pgfplotsset{compat=1.18}

\definecolor{lightblue}{RGB}{200, 230, 255}  
\definecolor{headerblue}{RGB}{150, 200, 255}

\newcounter{examplebox}

\makeatletter
\newcommand\blfootnote[1]{%
  \begingroup
  \renewcommand\thefootnote{}\footnote{#1}%
  \addtocounter{footnote}{-1}%
  \endgroup
}
\makeatother

\title{
    Unified Personalized Reward Model for Vision Generation
}

\author{
    Yibin Wang\textsuperscript{1,2}, 
    Yuhang Zang\textsuperscript{4$\dagger$},   
    Feng Han\textsuperscript{1,2},\\
    Jiazi Bu\textsuperscript{3,4},
    Yujie Zhou\textsuperscript{3,4},
    Cheng Jin\textsuperscript{1,2$\dagger$},
    Jiaqi Wang\textsuperscript{2$\dagger$}
}

\affiliation[1]{\mbox{Fudan University}}
\affiliation[2]{\mbox{Shanghai Innovation Institute}}

\affiliation[3]{\mbox{Shanghai Jiaotong University}}
\affiliation[4]{\mbox{Shanghai AI Lab}}

\newcommand{\ourmethod}{{\textsc {UnifiedReward-Flex}}\xspace}

\abstract{\input{sections/0abstract}
}

\checkdata[Website]{\url{https://codegoat24.github.io/UnifiedReward/flex}}

\begin{document}
\maketitle
\blfootnote{$^\dagger$Corresponding authors.}

\input{sections/1introduction}
\input{sections/2related}

\input{sections/3method}

\input{sections/4exp}

\input{sections/5conclusion}
\clearpage

\bibliographystyle{unsrtnat}
\bibliography{reference}

\clearpage

\beginappendix

\input{sections/7appendix}

\end{document}

%% file: sections/0abstract.tex
Recent advancements in multimodal reward models (RMs) have significantly propelled the development of visual generation. Existing frameworks typically adopt Bradley–Terry-style preference modeling or leverage generative vision-language models (VLMs) as judges, and subsequently optimize visual generation models via reinforcement learning. However, current RMs suffer from inherent limitations: they often follow a ``one-size-fits-all'' paradigm that assumes a monolithic preference distribution or relies on fixed evaluation rubrics. As a result, they are insensitive to content-specific visual cues, leading to systematic misalignment with subjective and context-dependent human preferences. To this end, inspired by human assessment, we propose \textbf{\ourmethod}, a unified personalized reward model for vision generation that couples reward modeling with flexible and context-adaptive reasoning. Specifically, given a prompt and the generated visual content, it first interprets the semantic intent and grounds on visual evidence, then dynamically constructs a hierarchical assessment by instantiating fine-grained criteria under both predefined and self-generated high-level dimensions. Our training pipeline follows a two-stage process: (1) we first distill structured, high-quality reasoning traces from advanced closed-source VLMs to bootstrap supervised fine-tuning (SFT), equipping the model with flexible and context-adaptive reasoning behaviors; (2) we then perform direct preference optimization (DPO) on carefully curated preference pairs to further strengthen reasoning fidelity and discriminative alignment. To validate the effectiveness, we integrate \ourmethod into the Group Relative Policy Optimization (GRPO) framework for image and video synthesis. Extensive results demonstrate its superiority: it consistently delivers more robust, context-aware reward signals than existing baselines and achieves substantial improvements across diverse image and video generation models.

%% file: sections/1introduction.tex
\section{Introduction}
With the rapid progress of multimodal reward models (RMs) \cite{unifiedreward,unifiedreward_think,LiFT,videoalign,zang2025internlm}, their potential in aligning vision generation models with human preferences has attracted growing attention. By converting subjective and high-level human judgments into learnable reward signals, RMs enable effective preference-driven post-training for vision generators. \cite{pref_grpo,zhou2025g2rpo,wu2025rewarddance,flowgrpo,xue2025dancegrpo,li2025mixgrpo}. Early reward modeling for visual generation typically uses fixed, discriminative scorers \cite{clip,hpsv2,aesthetics,pickscore} to assign scalar rewards. Later works shift to Bradley–Terry \cite{bradley1952rank} pairwise preference modeling to learn rewards from relative comparisons \cite{videoalign,ma2025hpsv3}. More recently, VLM-as-a-judge \cite{unifiedreward,unifiedreward_think,LiFT,videoscore2} has emerged, leveraging strong generative VLMs \cite{bai2025qwen2,li2024llava} to provide context-dependent evaluations for reinforcement learning. 

Despite their effectiveness, current RMs often follow a “one-size-fits-all” assessment paradigm:
\textbf{(1)} Fixed discriminative scorers (e.g., CLIP~\cite{clip}, PickScore~\cite{pickscore}) and Bradley--Terry preference models (e.g., VideoAlign~\cite{videoalign}, HPSv3~\cite{ma2025hpsv3}) typically learn a {single global} reward function that assigns a scalar score to each input (or induces preferences via score differences), implicitly assuming a monolithic preference distribution shared across diverse prompts and visual contents.
\textbf{(2)} VLM-as-a-judge approaches (e.g., UnifiedReward-Think~\cite{unifiedreward_think}) leverage generative VLMs to produce richer textual judgments, yet they often follow static evaluation rubrics with a fixed checklist of criteria.
As a result, their reward feedback remains insensitive to content-specific visual cues, which can misguide optimization and cause systematic misalignment with human preferences.

In this work, we posit that reliable reward assessment should be flexibly adapted to the prompt intent and visual content. For example, prompts with implicit narrative intent should emphasize storytelling consistency, subject relationships, and emotional tone (see Fig. \ref{fig:image_flex_case_1}), whereas motion-intensive videos with frequent physical contact demand explicit evaluation of action dynamics and physical plausibility (see Fig. \ref{fig:video_flex_case_1}). This behavior closely mirrors how humans assess visual generations: evaluators first interpret the prompt intent and the depicted content, then evaluate along a small set of common, task-specific high-level dimensions. Within each dimension, they selectively attend to the aspects most relevant to the given instance. When the scene exhibits additional characteristics, evaluators naturally introduce new high-level dimensions to capture these salient factors. In other words, human evaluation is inherently a content-adaptive, context-aware reasoning process, where both the evaluation criteria and their relative importance are dynamically adjusted to match the semantic intent and visual evidence.

As inspired, this work proposes \textbf{\ourmethod}, a unified personalized reward model for vision generation that couples reward modeling with context-adaptive reasoning to dynamically tailor evaluation criteria, which performs assessment in a hierarchical manner. 
Specifically, given a prompt and the generated visual content, it first interprets the semantic intent and extracts salient visual evidence, then composes a hierarchical evaluation plan by instantiating fine-grained sub-criteria under a few predefined coarse-grained dimensions. When the context demands additional considerations, it augments the hierarchy with new high-level dimensions and their corresponding sub-criteria. This dynamic criterion composition yields adaptive and informative reward feedback that aligns with the human evaluation process and remains robust across diverse prompts and visual content. Our training follows a two-stage pipeline: (1) We first distill structured, high-quality reasoning traces from advanced closed-source VLMs \cite{openai2025gpt52systemcard} to bootstrap supervised fine-tuning (SFT), endowing the model with content-aware criterion composition. (2) We then apply direct preference optimization (DPO) using preference pairs with human-grounded labels: given two sampled responses for the same input, if one reaches the correct conclusion while the other does not, the preference naturally favors the correct one; if both are correct, we further assign the preference based on the quality of their adaptive reasoning trajectories, prioritizing more flexible and context-grounded evaluation hierarchies. Empirically, this reasoning-aware preference alignment improves the reward model’s discriminative power, even among samples that are all correct.

Extensive experiments show that \ourmethod consistently outperforms strong reward model baselines on both image and video reward tasks, providing more robust and context-aware reward signals. To further validate its practical utility, we apply it as the reward signal for pairwise preference–based GRPO~\cite{pref_grpo} on multiple vision generation models. Specifically, we observe consistent improvements on text-to-image generators such as FLUX.1-dev and FLUX.2-klein-base, as well as on text-to-video generators including Wan2.1 and Wan2.2, yielding substantial quantitative and qualitative gains in downstream generation quality.

\textbf{Contributions.}
\textbf{(1)} We identify a key limitation of existing multimodal reward models, i.e., their ``one-size-fits-all'' evaluation paradigm, and propose \ourmethod, a unified personalized reward model for vision generation that dynamically composes evaluation hierarchies via content-aware reasoning.
\textbf{(2)} Our \ourmethod consistently outperforms strong reward model baselines on both image and video reward tasks, delivering more robust and context-aware reward signals.
\textbf{(3)} We further validate its practical utility by integrating it into pairwise preference reward-based GRPO for image and video synthesis, achieving substantial improvements in downstream generation quality both quantitatively and qualitatively.

\input{img/image_flex_case_1_}

\input{img/video_flex_case_1_}

%% file: img/image_flex_case_1_.tex
\begin{figure}[!ht]

    \centering
    \includegraphics[width=0.95\linewidth]{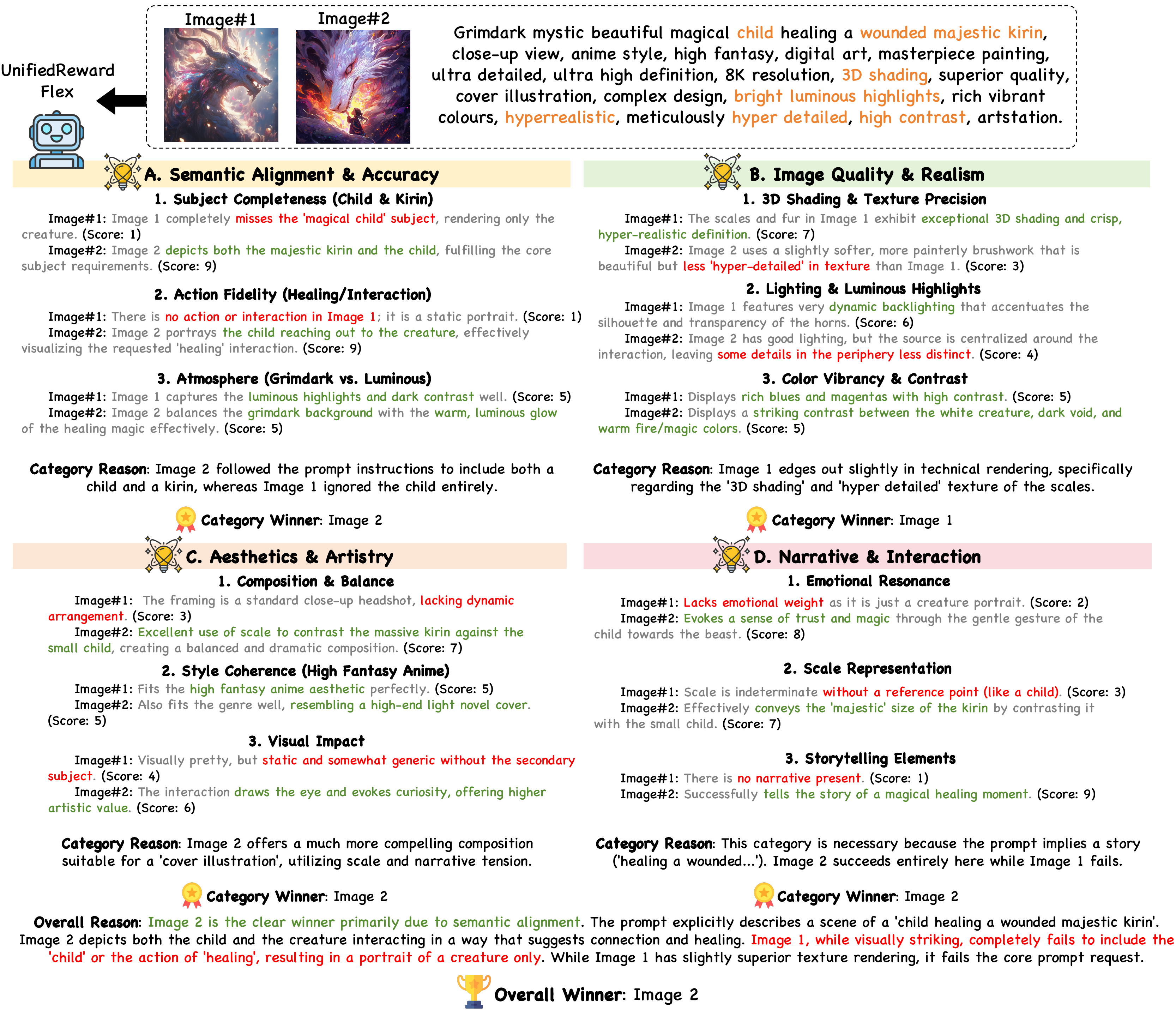}
    \caption{\textbf{Qualitative Result of \ourmethod on Image Generation Personalized Reward Reasoning.}}

    \label{fig:image_flex_case_1}

\end{figure}

%% file: img/video_flex_case_1_.tex
\begin{figure}[!ht]

    \centering
    \includegraphics[width=0.9\linewidth]{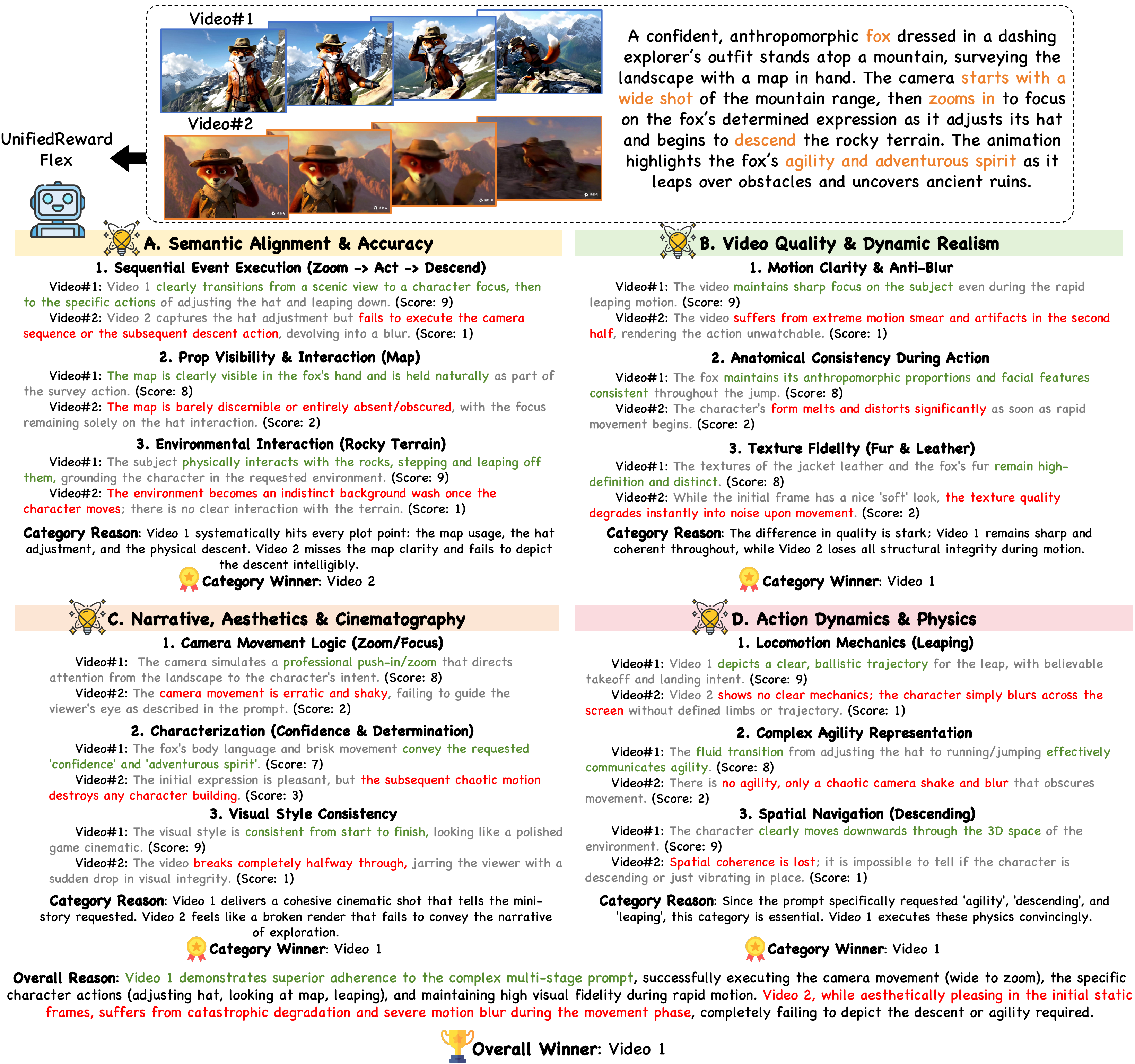}
    \caption{\textbf{Qualitative Result of \ourmethod on Video Generation Personalized Reward Reasoning.}}

    \label{fig:video_flex_case_1}

\end{figure}

%% file: sections/2related.tex
\section{Related Work}

\textbf{\textit{Multimodal Reward Models}} (RMs) are crucial in aligning vision generation models with human preferences. 
Early reward modeling typically relies on \textbf{fixed, discriminative scorers} that assign a scalar score to each generated sample~\cite{clip,aesthetics,pickscore,hpsv2}. 
These scorers are lightweight and easy to deploy, but they often behave as task-agnostic heuristics: the reward function is largely fixed across prompts and contents, making it difficult to reflect diverse evaluation focuses.
To better capture the relative nature of human judgment, later works shift from absolute scoring to pairwise preference modeling under the \textbf{Bradley-Terry pairwise preference modeling}~\cite{bradley1952rank}.
Instead of regressing a single target score, the RM is trained to assign higher reward to preferred samples within a comparison pair, which often yields more stable and better calibrated training signals for preference optimization~\cite{videoalign,ma2025hpsv3,zang2025internlm}. 
Nevertheless, Bradley--Terry based RMs still typically learn a single global scalar reward function shared across heterogeneous prompts and visual contents, implicitly assuming a monolithic preference distribution.
More recently, \textbf{VLM-as-a-judge} has emerged as a flexible alternative that leverages strong generative VLMs to directly assess and compare candidates~\cite{unifiedreward,unifiedreward_think,LiFT,videoscore2,wang2025vr,wang2025llava,xiong2025llava,xiong2025multi}. 
Compared to discriminative scorers, these judge models can provide richer evaluative feedback, potentially incorporating multi-aspect reasoning and content-dependent judgments. 
However, existing VLM-judge approaches commonly follow static evaluation rubrics or fixed checklists of criteria, which limits their ability to dynamically tailor evaluation priorities to the semantic intent and visual evidence of each input.
To this end, this work proposes \textbf{\ourmethod}, a unified personalized reward model for vision generation that couples reward modeling with content-aware reasoning. 
Unlike prior RMs that apply a fixed global scoring function or static rubrics, \ourmethod dynamically constructs a context-adaptive hierarchical assessment by self-instantiating fine-grained criteria based on given prompt intent and visual evidence.

\noindent\textbf{\textit{Reinforcement Learning for Vision Generation}} has evolved rapidly in recent years. 
Early attempts either fine-tuned models with scalar reward supervision~\cite{clark2023directly,prabhudesai2024video} or adopted reward-weighted regression to exploit reward feedback in a more stable manner~\cite{lee2023aligning,LiFT,videoalign}. 
Subsequent work drew inspiration from Proximal Policy Optimization (PPO)~\cite{schulman2017proximal} and incorporated policy-gradient updates into diffusion-based generators, showing promising improvements in sample quality~\cite{black2023training,fan2023reinforcement,miao2024training}. 
Despite their effectiveness, these methods are often computationally demanding and sensitive to hyperparameter tuning.
To improve training efficiency, a growing line of research explores preference optimization with supervised objectives, such as direct preference optimization (DPO), which directly leverages human preference data and avoids expensive on-policy rollouts~\cite{rafailov2023direct,wallace2024diffusion,yang2024using,liu2025videodpo}. 
More recently, Group Relative Policy Optimization (GPRO)~\cite{guo2025deepseek} has emerged as a promising objective for preference optimization in complex reasoning tasks. \cite{flowgrpo,xue2025dancegrpo} extend GRPO to flow matching models by reformulating the ODE sampling process as an equivalent SDE, enabling diverse sampling while performing preference-driven optimization. ~\cite{li2025mixgrpo,zhou2025g2rpo,deng2026densegrpo} further improve efficiency with a sliding-window mechanism that localizes SDE sampling and GRPO updates within a window while retaining ODE sampling elsewhere. 
However, these approaches typically optimize pointwise scalar rewards and may suffer from reward hacking~\cite{wu2025rewarddance,flowgrpo}. 
Pref-GRPO~\cite{pref_grpo} reveals that reward hacking is largely driven by ``illusory advantages'' induced by unreliable reward signals, and stabilizes learning by replacing pointwise rewards with pairwise preference-based feedback.
In this work, we integrate our \ourmethod into the Pref-GRPO framework for both image and video synthesis to validate its effectiveness.

%% file: sections/3method.tex
\section{Method}
\subsection{Overview}
Multimodal reward models (RMs) serve as learned proxies for human visual preferences.
However, existing RMs often follow a ``one-size-fits-all'' paradigm: they either assume a monolithic preference distribution captured by a single global scoring function, or rely on fixed evaluation rubrics that apply the same criteria uniformly across inputs. 
As a result, their reward feedback is often insensitive to content-specific visual cues and prompt intent, leading to systematic misalignment with inherently subjective and context-dependent human preferences. To this end, we propose \textbf{\ourmethod}, a unified personalized reward model for vision generation that couples reward modeling with context-adaptive reasoning to dynamically tailor evaluation criteria. 

In this section, we first present our unified personalized reward modeling framework (Sec.~\ref{sec:reward_modeling}), including the design of our context-adaptive reasoning process (Sec.~\ref{sec:reward_design}) and a two-stage training pipeline: (i) reasoning distillation for SFT (Sec.~\ref{sec:sft}) and (ii) reasoning-aware preference alignment via DPO (Sec.~\ref{sec:dpo}). 
We then describe how to apply \ourmethod to reinforcement learning for vision generation (Sec.~\ref{sec:rl}), starting with the necessary preliminaries (Sec.~\ref{sec:preliminary}), and followed by our employed personalized multi-dimensional preference rewards for GRPO (Sec.~\ref{sec:multi_dim_pref_reward}).

\subsection{Unified Personalized Reward Modeling} \label{sec:reward_modeling}

\subsubsection{Context-adaptive Reasoning Process Design} \label{sec:reward_design}
Our goal is to mimic how humans evaluate generated visual content. In practice, for a given task, human evaluators typically assess outputs along a small set of common high-level dimensions (e.g., semantic alignment and perceptual quality), while instantiating more fine-grained factors depending on the prompt intent and the visual evidence under each dimensions. When additional considerations become salient (e.g., Narrative \& Interaction in Fig. \ref{fig:image_flex_case_1}), they naturally extend the evaluation dimensions to better capture what matters for the current instance.

Guided by this principle, we design a context-adaptive hierarchical reasoning process that starts from three predefined common dimensions as stable anchors (e.g., semantic alignment, visual quality, and aesthetics for image generation). Under each anchor, the model instantiates prompt-specific sub-dimensions for assessment, and it can further introduce new high-level dimensions (with corresponding sub-dimensions) when required by the context. For each high-level dimension, the model aggregates the evidence across its instantiated sub-dimensions to produce an overall comparative analysis and a dimension-level winner; it then combines the outcomes from all high-level dimensions to determine a final overall winner.

Specifically, as shown in Fig.~\ref{fig:video_flex_case_1}, \ourmethod derives prompt-relevant sub-dimensions under the given anchors to jointly assess semantic adherence, visual quality under motion, and cinematographic coherence of generated videos. Since this case emphasizes rich motion and physical interactions, the model further introduces an additional high-level dimension (i.e., Action Dynamics \& Physics) to explicitly evaluate temporal coherence and physical plausibility, which is then incorporated into the final overall decision.

By tailoring the evaluation hierarchy to the generation context, our model provides richer reward supervision for preference optimization, improving both intent satisfaction and content-critical quality factors.

\subsubsection{Stage I: Reasoning Distillation for SFT}
\label{sec:sft}
To bootstrap the model with context-adaptive assessment behaviors, we first distill structured reasoning traces from the powerful closed-source VLM \cite{openai2025gpt52systemcard} and use them for supervised fine-tuning (SFT).

\paragraph{Distillation data.}
Let $\mathcal{D}=\{x_i\}_{i=1}^{N}$ be a set of preference-evaluation instances, where each input
\begin{equation}
x_i=(p_i, v_i^{(0)}, v_i^{(1)})
\end{equation}
contains a text prompt $p_i$ and a pair of candidate visual generations $\big(v_i^{(0)}, v_i^{(1)}\big)$.
Given $x_i$, the closed-source teacher model $\mathcal{T}$ outputs a structured evaluation trace
\begin{equation}
y_i^{\mathcal{T}}=\mathcal{T}(x_i)=\big(\mathcal{H}_i,\; \mathcal{R}_i,\; \mathcal{W}_i\big),
\end{equation}
where $\mathcal{H}_i=\{(d_k,\mathcal{S}_{i,k})\}_{k=1}^{K_i}$ denotes the instantiated high-level dimensions
$d_k$ together with their prompt-specific sub-dimensions $\mathcal{S}_{i,k}$, 
$\mathcal{R}_i$ is the corresponding evidence-grounded reasoning trace,
and $\mathcal{W}_i=\big(\{w_{i,k}\}_{k=1}^{K_i},\, w_i\big)$ are winner labels. 
Here $w_{i,k}\in\{0,1\}$ denotes the winner under dimension $d_k$ and $w_i\in\{0,1\}$ denotes the overall winner.

\paragraph{SFT objective.}
We fine-tune the base model \cite{unifiedreward_think} with parameters $\theta$ to imitate the teacher outputs via conditional language modeling:
\begin{equation}
\mathcal{L}_{\mathrm{SFT}}(\theta)
=-\sum_{i=1}^{N}\log p_{\theta}\!\left(y_i^{\mathcal{T}} \mid x_i\right)
=-\sum_{i=1}^{N}\sum_{t=1}^{|y_i^{\mathcal{T}}|}
\log p_{\theta}\!\left(y_{i,t}^{\mathcal{T}} \mid x_i, y_{i,<t}^{\mathcal{T}}\right).
\end{equation}
This stage initializes the model to generate structured, context-adaptive evaluations for paired visual comparisons, providing a strong foundation for subsequent preference alignment.

\subsubsection{Stage II: Reasoning-Aware Preference Alignment via DPO} \label{sec:dpo}
Building on the SFT-initialized context-adaptive evaluations, we further optimize it for preference discrimination under human-grounded supervision. Specifically, we apply Direct Preference Optimization (DPO) to jointly align the final decision and the quality of the adaptive reasoning trajectory.

\paragraph{Preference pair construction.}
For each input $x_i=(p_i,v_i^{(0)},v_i^{(1)})$, we assume an annotated preference label $w_i^\star\in\{0,1\}$ provided by the dataset,
where $w_i^\star=0$ indicates $v_i^{(0)}$ is preferred and $w_i^\star=1$ indicates $v_i^{(1)}$ is preferred.
Starting from the SFT model, we sample two candidate structured evaluations
\begin{equation}
y_i^{(a)},\,y_i^{(b)} \sim \pi_{\theta}(\cdot \mid x_i),
\end{equation}
each containing an overall predicted winner $\hat{w}(y)\in\{0,1\}$ and a context-adaptive reasoning trace.
We define the correctness indicator
\begin{equation}
c\!\left(y_i^{(j)}\right)=\mathbb{I}\!\left[\hat{w}\!\left(y_i^{(j)}\right)=w_i^\star\right], \quad j\in\{a,b\}.
\end{equation}
If exactly one sample is correct, we set the preference to favor the correct one. 
If both samples are correct, we further rank them by the quality of their adaptive reasoning trajectories, preferring evaluations with more context-grounded and flexible hierarchies.
Concretely, when both samples are correct, we obtain a trajectory-level preference by querying a closed-source judge $\mathcal{T}_{\mathrm{judge}}$ to perform pairwise comparison and output the preferred reasoning trace, followed by human verification on the judged pairs. 
We denote the judged preference as
\begin{equation}
\ell_i^{\mathrm{traj}}=\mathcal{T}_{\mathrm{judge}}(x_i, y_i^{(a)}, y_i^{(b)})\in\{a,b\},
\end{equation}
where $\ell_i^{\mathrm{traj}}$ indicates which trace is preferred in terms of reasoning quality. 
We then construct the DPO pair $(y_i^+,y_i^-)$ as
\begin{equation}
(y_i^+,y_i^-)=
\begin{cases}
(y_i^{(a)},y_i^{(b)}), & c(y_i^{(a)})>c(y_i^{(b)})\\
(y_i^{(b)},y_i^{(a)}), & c(y_i^{(b)})>c(y_i^{(a)})\\
(y_i^{(\ell_i^{\mathrm{traj}})},\, y_i^{(\bar{\ell}_i^{\mathrm{traj}})}), & c(y_i^{(a)})=c(y_i^{(b)})=1,
\end{cases}
\end{equation}
where $\bar{\ell}_i^{\mathrm{traj}}\in\{a,b\}\setminus\{\ell_i^{\mathrm{traj}}\}$, and we discard the pair if $c(y_i^{(a)})=c(y_i^{(b)})=0$. 
This construction aligns not only the final preference decision but also the quality of the underlying context-adaptive reasoning trajectory.

\paragraph{DPO objective.}
Given the resulting preference dataset $\mathcal{P}=\{(x_i,y_i^+,y_i^-)\}$, we optimize $\pi_\theta$ with the standard DPO loss using a frozen reference policy $\pi_{\mathrm{ref}}$ (the SFT model):
\begin{equation}
\mathcal{L}_{\mathrm{DPO}}(\theta)=
-\mathbb{E}_{(x,y^+,y^-)\sim\mathcal{P}}
\left[
\log \sigma\!\Big(
\beta_{dpo}\big(
\log \pi_\theta(y^+\!\mid x)-\log \pi_\theta(y^-\!\mid x)
-\log \pi_{\mathrm{ref}}(y^+\!\mid x)+\log \pi_{\mathrm{ref}}(y^-\!\mid x)
\big)
\Big)
\right],
\end{equation}
where $\beta$ controls the strength of preference optimization. By directly increasing the likelihood of preferred structured evaluations, this stage sharpens the reward model's discriminative ability while reinforcing its context-adaptive reasoning behavior. Empirically, such reasoning-aware preference alignment improves the reward model's discriminative power, even among samples that are all correct.

\subsection{Reinforcement Learning for Vision Generation} \label{sec:rl}
\subsubsection{Preliminaries}
\label{sec:preliminary}
\textbf{Flow Matching GRPO.}
We briefly review GRPO~\citep{guo2025deepseek} in the context of flow-matching models~\citep{flowgrpo,xue2025dancegrpo}. 
Given a prompt $c$, a flow-based generator produces an image/video $x_0$ by iteratively refining a noisy sample from $t=T$ to $t=0$. For each sample, we consider terminal reward supervision, where the reward is provided at the end of the trajectory as $R(x_0,c)$.

\paragraph{Group-relative advantage.}
Given a prompt $c$, GRPO samples a group of $G$ generations $\{x_0^i\}_{i=1}^{G}$ and evaluates them with a reward model to obtain $\{R(x_0^i,c)\}_{i=1}^{G}$.
To stabilize learning under noisy rewards, GRPO constructs a prompt-wise standardized advantage:
\begin{equation}
\hat{A}^i \;=\;
\frac{R(x_0^i,c)-\mu_c}{\sigma_c},
\qquad
\mu_c=\mathrm{mean}\big(\{R(x_0^j,c)\}_{j=1}^{G}\big),\quad
\sigma_c=\mathrm{std}\big(\{R(x_0^j,c)\}_{j=1}^{G}\big).
\label{eq:grpo_adv}
\end{equation}
This relative advantage encourages the policy to increase the likelihood of higher-quality generations within the same prompt group, rather than chasing absolute reward values.

\paragraph{GRPO objective.}
Let $\pi_\theta$ denote the sampling policy induced by the generator, and $\pi_{\theta_{\text{old}}}$ be the behavior policy used to collect trajectories.
For the $i$-th trajectory at step $t$, the importance ratio is
\begin{equation}
r_t^i(\theta)=\frac{\pi_\theta(x_{t-1}^i\mid x_t^i,c)}{\pi_{\theta_{\text{old}}}(x_{t-1}^i\mid x_t^i,c)}.
\end{equation}
GRPO then maximizes a clipped surrogate objective with KL regularization to a reference policy $\pi_{\text{ref}}$:
\begin{equation}
\mathcal{J}_{\text{GRPO}}(\theta)
=
\mathbb{E}_{c}\!\left[
\frac{1}{G}\sum_{i=1}^{G}\frac{1}{T}\sum_{t=0}^{T-1}
\min\!\Big(r_t^i(\theta)\hat{A}^i,\;
\text{clip}\big(r_t^i(\theta),1-\eta,1+\eta\big)\hat{A}^i\Big)
\;-\;
\beta_{kl}\, D_{\text{KL}}(\pi_\theta \,\|\, \pi_{\text{ref}})
\right],
\label{eq:grpo_obj}
\end{equation}
where $\eta$ is the clipping threshold and $\beta_{kl}$ controls the strength of regularization. 
This formulation provides a principled and stable way to post-train vision generators with learned reward signals.

\noindent\textbf{Pairwise Preference Reward-based GRPO} (Pref-GRPO)~\cite{pref_grpo}
replaces absolute reward scores with relative preference judgments, which better align with the common practice of evaluating visual generations via pairwise comparisons.
Instead of assigning an independent scalar reward to each sample, Pref-GRPO derives terminal rewards from comparative outcomes among a group of generated samples.

Given a prompt $c$, a group of $G$ images (or videos) $\{x_0^i\}_{i=1}^G$ is sampled from the current policy $\pi_\theta$.
For each unordered pair $(x_0^i, x_0^j)$, a preference model determines which sample is preferred, denoted by $x_0^i \succ x_0^j$.
Based on all pairwise comparisons within the group, Pref-GRPO defines a relative preference reward for each sample as its normalized win rate:
\begin{equation}
R(x_0^i, c)
\;=\;
\frac{1}{G-1} \sum_{j \neq i} \mathbb{1}\big(x_0^i \succ x_0^j\big),
\end{equation}
which reflects how often sample $i$ is preferred over other candidates under the same prompt.

These preference-derived rewards are then directly plugged into the standard GRPO framework.
In particular, the group-relative advantage $\hat{A}^i$ is computed using Eq.~\eqref{eq:grpo_adv}, and the policy is optimized with the same clipped surrogate objective in Eq.~\eqref{eq:grpo_obj}.

\subsubsection{Personalized Multi-Dimensional Preference Rewards}
\label{sec:multi_dim_pref_reward}

In this work, we integrate our flexible, context-adaptive evaluations from \ourmethod into Pref-GRPO to provide personalized, multi-dimensional rewards. Specifically, given a prompt $c$, the policy $\pi_\theta$ samples a group of $G$ candidates $\{x_0^i\}_{i=1}^{G}$, and for each unordered pair $(x_0^i, x_0^j)$, the reward model produces pairwise preference judgments along $D$ predefined anchor dimensions:
\[
x_0^i \succ_d x_0^j, \quad d=1,\dots,D.
\]

For each candidate, we compute the dimension-wise win rates and their average over the anchor dimensions:
\[
\bar{R}_{\mathrm{dim}}(x_0^i, c) = \frac{1}{D} \sum_{d=1}^{D} R_d(x_0^i, c), \quad
R_d(x_0^i, c) = \frac{1}{G-1} \sum_{j \neq i} \mathbb{1}\big(x_0^i \succ_d x_0^j\big).
\]

To account for dynamic, personalized high-level dimensions that may not appear consistently in each pairwise comparison, we also compute the overall win rate:
\[
R_{\mathrm{overall}}(x_0^i, c) = \frac{1}{G-1}\sum_{j\neq i} \mathbb{1}\big(x_0^i \succ x_0^j\big).
\]

The group-relative advantages are then computed separately for the averaged dimension-wise win rate and the overall win rate:
\[
\hat{A}^i_{\mathrm{dim}} = \frac{\bar{R}_{\mathrm{dim}}(x_0^i,c)-\mu_{\mathrm{dim}}}{\sigma_{\mathrm{dim}}}, \quad
\hat{A}^i_{\mathrm{overall}} = \frac{R_{\mathrm{overall}}(x_0^i,c)-\mu_{\mathrm{overall}}}{\sigma_{\mathrm{overall}}},
\]
where $\mu$ and $\sigma$ are the mean and standard deviation within the prompt group.

Finally, the combined advantage used in GRPO is
\[
\hat{A}^i = \alpha\, \hat{A}^i_{\mathrm{dim}} + (1-\alpha)\, \hat{A}^i_{\mathrm{overall}},
\]
where \(\alpha\) controls the relative contributions of the averaged dimension-wise advantage and the overall advantage, respectively.

This procedure ensures that both the fine-grained anchor preferences and the holistic, context-adaptive evaluation contribute to the policy update, while remaining compatible with the training objective in Eq.~\eqref{eq:grpo_obj}.

%% file: sections/4exp.tex
\section{Experiment}
\input{table/reward_compare}
\input{table/unigenbench}
\input{table/image_eval}

\subsection{Implementation Details}
\subsubsection{\ourmethod}
\textbf{Datasets.}
For \textit{\textbf{image generation}}, we sample 50K image preference pairs from HPDv3~\citep{ma2025hpsv3}.
For \textit{\textbf{video generation}}, we combine two human preference datasets: Text2Video-Human Preferences (15K), provided by Rapidata, and VideoFeedback2~\citep{videoscore2}, from which we preprocess and construct an additional 35K video preference pairs.
In the SFT stage, we distill reward reasoning traces for both images and videos from GPT-5.2 \cite{openai2025gpt52systemcard}, using 45K image pairs and 45K video pairs to construct our \textit{\textbf{UnifiedReward-Flex-SFT-90K}} dataset.
Training samples are randomly drawn from the collected datasets, while the remaining data are reserved for the subsequent DPO stage. In the DPO stage, we use non-greedy decoding with temperature 0.7 and sample two reasoning traces per prompt.
If both samples yield the correct preference decision, we further use GPT-5.2 to perform pairwise comparison of their reasoning trajectories and select the higher-quality trace as the preferred response for constructing DPO training pairs.

\noindent\textbf{Reward model.} We adopt UnifiedReward-Think-qwen3vl~\citep{unifiedreward_think} as the base reward model, which supports long-chain reasoning for both visual perception and generation.
Building on its strong prior knowledge, we further steer it toward flexible, context-adaptive reward reasoning. To evaluate robustness across model scales, \ourmethod is trained with backbone sizes ranging from 2B$\sim$32B. The 8B variant is used as the default reward model in all vision generation GRPO experiments, while other scales are included to analyze the effect of model capacity on reward assessment performance.

\noindent\textbf{Training details.} 
Our training is conducted with a batch size of 2 and 2 gradient accumulation steps, using a learning rate of $2.5\times10^{-6}$ and a warm-up ratio of 0.1. All experiments are performed on 32 NVIDIA H200 GPUs.
For the DPO stage, we set $\beta_{{dpo}}$ to 0.1. 

\noindent\textbf{Evaluation.} We evaluate image reward models on GenAI-Bench-Image \cite{jiang2024genai} and Multimodal RewardBench 2 (MMRB2) \cite{hu2025multimodalrewardbench2}, and video reward models on GenAI-Bench-Video \cite{jiang2024genai} and MJ-Bench-Video \cite{mjvideo}.
For pairwise preference-based reward models, we randomly permute the order of the two candidates at evaluation time.

\subsubsection{Reinforcement Learning for Vision Generation}
\textbf{Text-to-Image Generation.}
\textit{\textbf{Training.}} We perform GRPO on FLUX.1-dev~\cite{flux} using training prompts from UniGenBench++~\cite{unigenbench++}. Training is conducted on 32 NVIDIA H200 GPUs with 15 sampling steps and 9 rollouts per prompt from the same initial noise, using 3 gradient accumulation steps and a learning rate of $3\times10^{-6}$. We set $\beta_{\mathrm{KL}}=0$.
\textit{\textbf{Evaluation.}} For inference, we use 30 sampling steps and a classifier-free guidance scale of 3.5, following the official configuration. We evaluate in-domain performance on UniGenBench++. For out-of-domain evaluation, we measure semantic consistency on GenEval~\cite{ghosh2023geneval} and T2I-CompBench~\cite{t2i-compbench}, and assess image quality using UnifiedReward~\cite{unifiedreward}, Pickscore \cite{pickscore}, and the aesthetic predictor~\cite{aesthetics}.

\noindent\textbf{Text-to-Video Generation.}
\textit{\textbf{Training.}} We perform GRPO on Wan2.1-T2V-14B~\citep{wan2} using the training prompts provided by~\citep{xue2025dancegrpo}. Training is conducted on 32 NVIDIA H200 GPUs with LoRA rank 64 and alpha 128, using 20 sampling steps and 6 rollouts per prompt from the same initial noise. We train with videos at $240\times416$ resolution with 33 frames, 2 gradient accumulation steps, and a learning rate of $3\times10^{-5}$. We set $\beta_{\mathrm{KL}}=0.004$.
\textit{\textbf{Evaluation.}} For inference, we use 30 sampling steps and a classifier-free guidance scale of 5, generating videos at $480\times832$ resolution with 33 frames, and evaluating the performance on VBench~\citep{huang2023vbench}.

To assess \ourmethod's robustness across different generators, we additionally perform GRPO on FLUX.2-klein-base-9B \cite{flux2} and Wan2.2-T2V-A14B \cite{wan2} under the same experimental settings as above.

\input{img/image_compare1}

\subsection{Results}

\subsubsection{Reward Model Comparison}
We compare our \ourmethod against representative baselines on both image and video preference-evaluation benchmarks. 
Fixed scorers (e.g., HPSv2 \cite{hpsv2}, PickScore \cite{pickscore}) and Bradley--Terry preference models (e.g., HPSv3 \cite{ma2025hpsv3}, VideoReward \cite{videoalign}) provide a single global signal that is insensitive to prompt-specific requirements. 
Besides, UnifiedReward-Think~\cite{unifiedreward_think} is a strong VLM-as-a-judge baseline that performs multi-modal assessment with long-chain reasoning, but still follows a fixed checklist of evaluation criteria.
In contrast, \ourmethod dynamically instantiates fine-grained criteria conditioned on the prompt and visual evidence, yielding more reliable pairwise decisions. \textbf{Quantitatively}, as shown in Tab. \ref{tab:reward_benchmark}, \ourmethod achieves the best performance across all benchmarks; notably, it improves over UnifiedReward-Think by +3.2 points on MMRB2 and +2.2 points on GenAI-Bench-Video, highlighting the benefit of context-adaptive criterion composition beyond long-chain reasoning alone.
\textbf{Qualitative} results are provided in Figs. \ref{fig:image_flex_case_1} and \ref{fig:video_flex_case_1}. For example, Fig.~\ref{fig:image_flex_case_1} highlights our hierarchical, context-adaptive evaluation in a story-implied prompt (``a child healing a wounded kirin''). 
Starting from common anchor dimensions (semantic alignment, visual quality, and aesthetics), our model instantiates prompt-specific sub-criteria to check core requirements such as subject completeness and style coherence. 
Crucially, because the prompt implicitly describes a narrative moment rather than a static portrait, the model further augments the evaluation hierarchy with an additional high-level dimension, i.e., ``Narrative \& Interaction'', to judge whether the generation conveys a coherent healing scene with meaningful entity relations. 
This adaptive criterion composition prevents the evaluation from being dominated by surface-level rendering quality alone, and yields preference judgments that better reflect prompt-critical intent, providing more informative reward supervision for downstream optimization.
\input{table/vbench}

\input{img/video_compare1}
\subsubsection{Text-to-Image GRPO}
\noindent\textbf{Quantitatively},
on UniGenBench (Tab.~\ref{tab:unigenbench}), \ourmethod improves the overall semantic consistency (+14.56) over the base model, and also surpasses the strong VLM-as-a-judge baseline UnifiedReward-Think (+5.06).
The improvements are broad-based across challenging dimensions that require compositional and intent-aware evaluation, such as Compound and Logical Reasoning.
It also generalizes well to out-of-domain benchmarks (Tab.~\ref{tab:ood_compare}): achieves the best semantic consistency on T2I-CompBench \cite{t2i-compbench} and GenEval \cite{ghosh2023geneval}, while maintaining or improving image quality metrics (e.g., UnifiedReward \cite{unifiedreward}) compared to other reward baselines.
These results indicate that optimizing with our personalized reward does not merely overfit to in-domain prompts; instead, it promotes more robust alignment with diverse prompt intents and visual evidence. 

\noindent\textbf{Qualitatively}, Fig.~\ref{fig:qualitative_image_compare} further shows that our model better enforces prompt-critical constraints. 
In the ``Newton'' example (row~2), the prompt couples multiple requirements (a square falling apple and a circular shadow), which demands reasoning over attributes and geometry. 
Among the compared methods, only the generator optimized with our reward consistently satisfies this coupled constraint.

\subsubsection{Text-to-Video GRPO}
\textbf{Quantitatively}, Tab.~\ref{tab:vbench_compare} summarizes VBench results for Wan2.1-T2V-14B GRPO with different reward signals. 
\ourmethod yields clear improvements on both dynamic quality and compositional semantics. For example,
on the \textbf{\textit{quality}} side, it notably boosts the Dynamic Degree score (from 58.6 to 70.8), indicating that our reward provides more informative supervision for motion-intensive generations beyond static appearance. 
On the \textbf{\textit{semantic}} side, it substantially improves Spatial Relationship (72.6 to 80.8) and Color consistency (87.7 to 89.6), suggesting better prompt grounding in relational and attribute-level constraints. 
Compared with VideoReward and UnifiedReward-Think, while both baselines also achieve competitive results, their improvements are less pronounced; in particular, they incur drops on Dynamic Degree, indicating limited ability to encourage richer motion. This contrast further supports that our context-personalized evaluation provides more effective reward supervision for GRPO-based video post-training.

\noindent\textbf{Qualitatively}, Fig.~\ref{fig:qualitative_video_compare} shows that using \ourmethod as the reward for GRPO yields more coherent videos under motion- and interaction-centric prompts.
In the upper-left example (``Two AI models fighting in Mortal Kombat''), the base Wan2.1-T2V-14B output is largely static, and GRPO with VideoReward or UnifiedReward-Think often further dampens the motion dynamics, resulting in reduced action amplitude and weaker interaction intensity across frames.
In contrast, \ourmethod-guided GRPO preserves stronger, continuous motion and clearer contact-driven progression, better matching the prompt’s intent.

\input{table/ablation_reward}


\subsection{Ablation Studies}

\subsubsection{Effect of DPO Alignment}
As shown in Tab.~\ref{tab:reward_benchmark}, applying DPO yields consistent improvements, validating the role of preference alignment in sharpening preference discrimination.
Importantly, the gains persist even in the ``Both correct'' setting, where both sampled traces predict the correct final preference.
In this harder regime, DPO provides supervision on how the decision is reached: explicitly preferring higher-quality, more context-adaptive reasoning trajectories further improves discriminative accuracy by separating subtle quality differences that are invisible to decision-level correctness alone.
Together, these results motivate coupling personalized criterion composition with reasoning-aware preference alignment to build robust reward models for both image and video generation.

\subsubsection{Hyperparameter Analysis of $\alpha$}
We analyze the impact of \(\alpha\) in Tab. \ref{tab:ablation_reward}, which controls the balance between the averaged dimension-wise win rate (\(\bar{R}_{\mathrm{dim}}\)) and the overall win rate (\(R_{\mathrm{overall}}\)) in our reward model. When \(\alpha=0\), the model relies solely on the overall win rate, potentially overlooking finer, context-specific details, resulting in suboptimal performance. As \(\alpha\) increases, the model shifts focus towards dimension-wise rewards, capturing more context-adaptive nuances. However, when \(\alpha=1\), the model becomes overly focused on dimension-wise rewards, limiting its ability to assess broader, global preferences. Our results show that \(\alpha=0.7\) strikes an optimal balance, combining context-aware reasoning with global quality assessment, leading to the best performance across both image and video generation tasks. This further demonstrates the effectiveness of our personalized multi-dimensional rewards in enhancing the model’s preference alignment.

\clearpage

\input{table/flex_scale}
\subsection{Discussion}

\subsubsection{Reward Assessment Performance across Model Scales}
Tab.~\ref{tab:reward_scale} reports the assessment performance of \ourmethod across different model scales. We observe a consistent improvement as model capacity increases, on both image and video generation benchmarks, indicating that \ourmethod benefits from additional representational and reasoning capacity while preserving stable behavior across scales.
Notably, the gains are smooth rather than abrupt: smaller models already achieve competitive performance, while larger models provide incremental improvements. This suggests that the core advantage of our model does not stem solely from model size, but from its context-adaptive evaluation mechanism, which remains effective even under limited capacity.

\input{img/klein_qualitative_results1}
\input{table/klein_unigenbench}
\input{table/klein_image_eval}

\input{table/vbench_wan22}
\input{img/wan22_qualitative_results1}

\subsubsection{Training Efficiency}
Tab.~\ref{tab:ablation_reward} compares the per-step training time across different reward models.
Fixed scorers and Bradley-Terry–style models (e.g., PickScore, HPSv3) are the most efficient, as they directly output scalar rewards without explicit reasoning.
However, this efficiency comes at the cost of coarse and less comprehensive reward signals with limited expressiveness.
Reasoning-based VLMs introduce additional overhead. Specifically,
UnifiedReward-Think is noticeably slower, since it performs long-chain reasoning under a fixed evaluation rubric to produce judgments. Our \ourmethod further increases the computation cost, as it goes beyond fixed criteria and conducts context-personalized reasoning: it dynamically instantiates fine-grained sub-criteria and, when necessary, introduces new high-level dimensions conditioned on the prompt intent and visual evidence, rather than applying a static checklist.
Despite this extra compute, \ourmethod delivers substantially stronger reward accuracy (Tab. \ref{tab:reward_benchmark}) and consistently larger gains in GRPO for both image and video generation (Tabs. \ref{tab:ood_compare} and \ref{tab:vbench_compare}).
These results indicate that the added training cost is a reasonable and practical trade-off for improved alignment quality and generation performance.

\input{table/efficiency}
\input{img/image_train_step1}
\input{img/clip_curve1}

\input{img/video_train_step1}

\subsubsection{Training Progress Visualization} 
\textbf{{Quantitatively}.}
Fig.~\ref{fig:clip_curve} visualizes the training dynamics when using CLIP score as a proxy metric to monitor semantic alignment during GRPO with \ourmethod.
Across both text-to-image (FLUX.1-dev) and text-to-video (Wan2.1-T2V-14B) settings, the CLIP score exhibits a clear and consistent upward trend over training steps, indicating steadily improved prompt-image/video semantic consistency.
The similarly stable increase in both domains suggests that our context-adaptive reward provides reliable optimization signals, reinforcing semantic understanding for both static image synthesis and temporal video generation. 

\noindent\textbf{{Qualitatively}.}
Figs. \ref{fig:image_train_step} and \ref{fig:video_train_step} show a consistent visual improvement trend during GRPO when using \ourmethod as the reward. For image generation (Fig.~\ref{fig:image_train_step}), later checkpoints better enforce prompt-critical constraints and compositional intent (e.g., the chest engraving becomes clearer and more faithful), indicating stronger semantic grounding beyond surface aesthetics. For video generation (Fig.~\ref{fig:video_train_step}), the origami-dancer example evolves from limited, less coordinated motion to smoother, more coherent action progression with improved temporal consistency across frames. Overall, these training-time visualizations corroborate that \ourmethod provides reliable, context-adaptive reward guidance that steadily strengthens both semantic adherence and temporal coherence during training.

\subsubsection{Robustness across Different Vision Generators}
To evaluate the robustness of \ourmethod, we apply it as the reward signal for GRPO across multiple image and video generators with diverse architectures and capabilities.
\textbf{Text-to-Video Generation (Tab.~\ref{tab:vbench_compare_wan22}).}
Across both Wan2.1-T2V-14B and the stronger Wan2.2-T2V-A14B, \ourmethod consistently improves motion- and interaction-related metrics, including {Dynamic Degree}, {Motion Smoothness}, and {Human Action}.
These gains indicate that the proposed reward encourages richer and more sustained motion patterns, rather than static or over-smoothed trajectories often induced by less expressive rewards.
Importantly, improvements are also observed on semantic metrics such as {Spatial Relationship} and {Multiple Objects}, suggesting that enhanced motion fidelity does not come at the expense of semantic structure, but instead reinforces both jointly.
\textbf{Text-to-Image Generation (Tabs. \ref{tab:unigenbench_klein} and \ref{tab:ood_compare_klein}).}
Similar robustness is observed across FLUX.1-dev and the stronger FLUX.2-klein-base-9B.
\ourmethod consistently improves semantic consistency across multiple benchmarks, while also yielding gains in image quality under different evaluation models.
This indicates that the learned reward generalizes well beyond a specific generator or evaluation setup.
\textbf{Overall}, the consistent gains suggest that \ourmethod does not rely on the specific generator. Instead, by dynamically instantiating evaluation criteria conditioned on prompt intent and visual evidence, it produces preference signals that remain effective across diverse output distributions.
This generator-agnostic behavior highlights the robustness of context-adaptive reward modeling for vision generation post-training.

%% file: table/reward_compare.tex
\begin{table}[t]
\centering
\small
\renewcommand{\arraystretch}{1.1}
\caption{\textbf{Image and Video Generation Assessment Comparison.}}
\setlength{\tabcolsep}{4pt}
\begin{tabular}{c c c | c c c}
\toprule
\multirow{2}{*}{\textbf{Method}} 
& \multicolumn{2}{c}{\textbf{Image Generation}} 
& \multirow{2}{*}{\textbf{Method}} 
& \multicolumn{2}{c}{\textbf{Video Generation}}  \\
\cmidrule(lr){2-3}
\cmidrule(lr){5-6}
& \textbf{GenAI-Bench} & \multicolumn{1}{c}{\textbf{MMRB2}} & & \textbf{GenAI-Bench} & \textbf{MJBench} \\
\hline
HPSv2          & 68.8 & 55.0 & LiFT           & 60.1 & 51.0 \\
PickScore      & 70.0 & 57.6 & VideoScore     & 70.6 & 62.8 \\
HPSv3          & 70.9 & 58.5 & VideoReward    & 73.1 & 63.4 \\
UnifiedReward  & 71.5 & 60.0 & UnifiedReward  & 76.8 & 68.8 \\
UnifiedReward-Think  
               & \underline{72.3} & 66.0 & UnifiedReward-Think  
               & 80.3 & \underline{70.9} \\
\hline
\textbf{Ours} w/o DPO & 71.5 & 67.5  & \textbf{Ours} w/o DPO
               & 79.4 & 69.1\\
\textbf{Ours} w/o DPO (Both correct) & 72.0 & \underline{68.4}  & \textbf{Ours} w/o DPO (Both correct)
               & \underline{80.6} & 70.3\\
\hline
\textbf{Ours}  & \textbf{73.4} & \textbf{69.2}  & \textbf{Ours}  
               & \textbf{82.5} & \textbf{72.0}\\
\bottomrule
\end{tabular}
\label{tab:reward_benchmark}
\end{table}

%% file: table/unigenbench.tex
\begin{table*}[t]
\centering
\scriptsize
\setlength{\tabcolsep}{4pt}
\renewcommand{\arraystretch}{1.2}
\caption{\textbf{In-domain Semantic Consistency Comparison on UniGenBench}. ``UniGenBench-EvalModel-qwen3vl-32b-v1'' is used as the VLM for evaluation.} 
\begin{tabular}{lc|cccccccccc}
\toprule
\textbf{Model} & \textbf{Overall} & Style & World Know. & Attribute & Action & Relation. & Compound & Grammar & Logic.Reason. & Layout & Text \\
\midrule
FLUX.1-dev        &59.39   & 85.10 & 85.92 & 65.28 & 61.41 & 64.97 & 43.56 & 60.16 & 24.77 & 70.52 & 32.18 \\
w/ HPSv2 & 57.77   & 77.90 & 87.03 & 65.92 & 57.41 & 65.86 & 44.46 & 55.75 & 29.09 & 64.93 & 29.31 \\
w/ HPSv3  &57.98  & 79.40 & 90.03 & 66.24 & 57.89 & 63.58 & 39.82 & 58.82 & 24.09 & 67.16 & 32.76 \\
w/ PickScore  &58.63  & 79.70 & 87.03 & 64.42 & 61.12 & 67.64 & 47.42 & 58.02 & 27.50 & 67.54 & 25.86 \\
w/ UnifiedReward  &60.87  & 83.50 & 87.97 & 66.13 & 63.88 & 68.65 & 46.52 & 58.69 & 24.32 & 71.08 & 37.93 \\
w/ UnifiedReward-Think  & \underline{68.89}  & \underline{88.00} & \textbf{91.77} & \underline{77.99} & \underline{69.20} & \underline{75.13} & \underline{61.47} & \underline{61.63} & \underline{41.36} & \underline{77.24} & \underline{45.11} \\
\hline
\textbf{w/ UnifiedReward-Flex} & \textbf{73.95} &\textbf{90.30}   & \underline{89.87} & \textbf{79.38} & \textbf{73.38} & \textbf{78.55}& \textbf{69.46} & \textbf{63.10} & \textbf{46.59} & \textbf{79.66} & \textbf{59.20}  \\
\bottomrule
\label{tab:unigenbench}

\end{tabular} 

\vspace{-0.1cm}
\end{table*}

%% file: table/image_eval.tex
\begin{table}[t]
\centering
\scriptsize
\setlength{\tabcolsep}{9pt}
\caption{\textbf{Out-of-Domain Semantic Consistency and Image Quality Evaluations}. The best results are in \textbf{bold}, and the second best are \underline{underlined}.}
\begin{tabular}{lccccccccccc}

\toprule
\multirow{2}{*}{\textbf{Model}}& \multicolumn{4}{c}{\textbf{Semantic Consistency}} & \multicolumn{3}{c}{\textbf{Image Quality}}\\
 \cmidrule(lr){2-5} \cmidrule(lr){6-8}
 & \textbf{UniGenBench}& \textbf{T2I-CompBench} &\textbf{GenEval} & \textbf{CLIP} & \textbf{PickScore} & \textbf{UnifiedReward} & \textbf{Aesthetic} \\
\midrule
FLUX.1-dev   & 59.39 &48.57 &  62.18  & 34.40 & 22.70 & 3.07 &  6.13 \\

w/ HPSv2   & 57.77 &44.84 &  58.43  & 33.35 & 23.12 & 3.10 &  6.23 \\
w/ HPSv3 & 57.98  & 46.46 &  61.14  & 34.12 & 23.26 & 3.14 &  6.37 \\
w/ PickScore & 58.63  & 45.92 &  58.76  & 33.61 & \textbf{23.78} & 3.12 &  6.42 \\
w/ UnifiedReward & 60.87  & 49.13 &  66.25  & 34.43 & 23.31 & 3.19 &  6.44 \\
w/ UnifiedReward-Think & \underline{68.89}  & \underline{50.10} &  \underline{68.20}  & \underline{35.85} & 23.38 & \underline{3.27} &  \underline{6.53} \\
\hline

\textbf{w/ UnifiedReward-Flex} &\textbf{73.95} & \textbf{51.37} &  \textbf{69.62}  & \textbf{36.25} & \underline{23.42} & \textbf{3.31}& \textbf{6.56}\\

\bottomrule
\label{tab:ood_compare}
\vspace{-0.2cm}
\end{tabular}

\end{table}

%% file: img/image_compare1.tex
\begin{figure}[!t]

    \centering
    \includegraphics[width=1\linewidth]{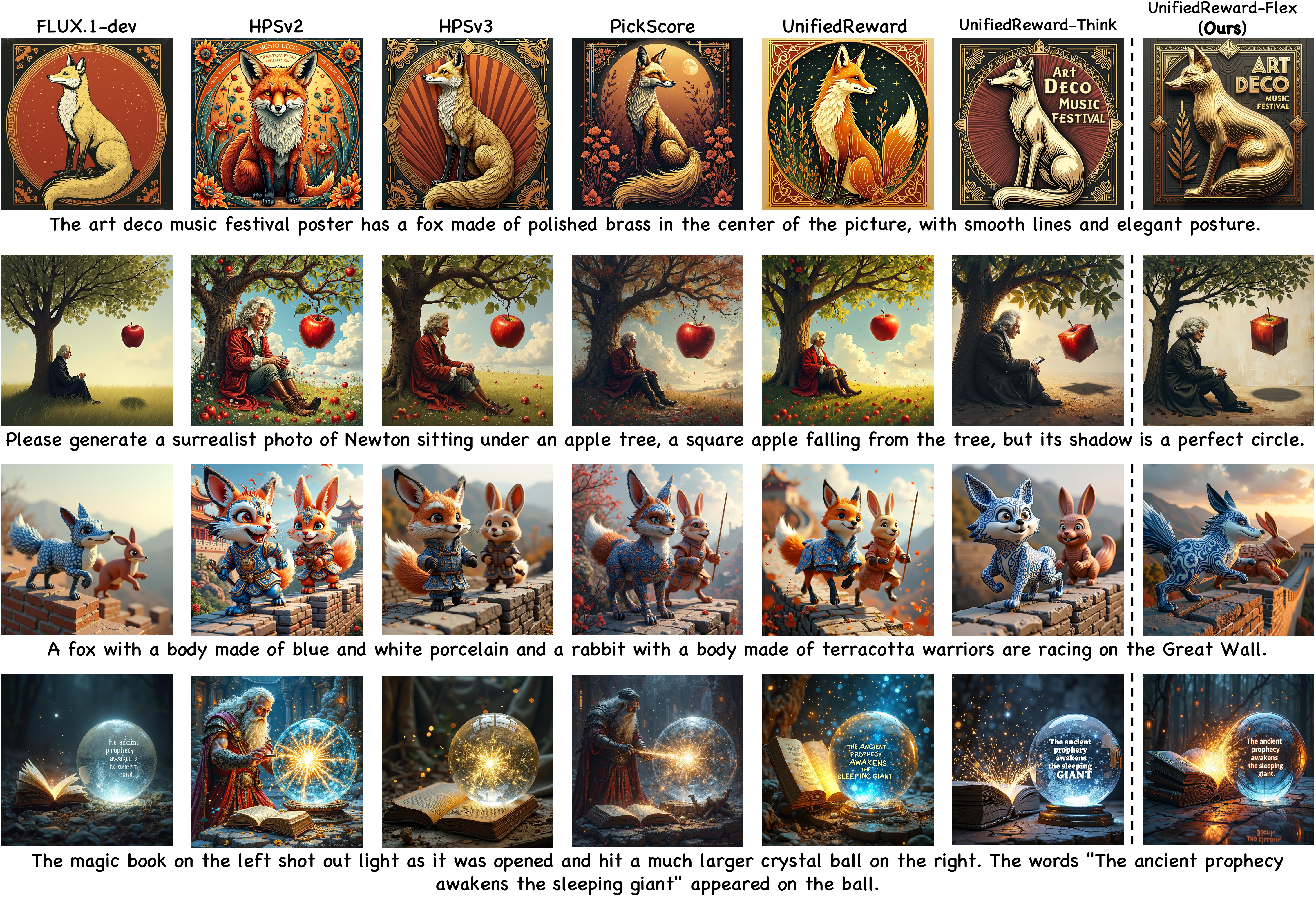}
    \caption{\textbf{Qualitative Comparison on Text-to-Image GRPO (FLUX.1-dev).}
    }

    \label{fig:qualitative_image_compare}

\end{figure}

%% file: table/vbench.tex
\begin{table*}[!th]
\setlength\tabcolsep{6pt}
\centering
\scriptsize
\caption{\textbf{Quantitative results on VBench}. The first seven metrics correspond to the \textit{Quality} type, while the remaining correspond to the \textit{Semantic} type.}

\begin{tabular}{ccccccccccc}

		\toprule
         Models &
	\makecell[c]{\textbf{Subject}\\ \textbf{Consistency}}
      & \makecell[c]{\textbf{Background}\\ \textbf{Consistency}}
  &\makecell[c]{\textbf{Aesthetic} \\\textbf{Quality}} &\makecell[c]{\textbf{Imaging} \\\textbf{Quality}}

&\makecell[c]{\textbf{Temporal} \\\textbf{Flickering}}
&\makecell[c]{\textbf{Motion} \\\textbf{Smoothness}}
& \makecell[c]{\textbf{Dynamic} \\\textbf{Degree}}
& \makecell[c]{\textbf{Human} \\\textbf{Action}}

       \\
		\midrule

 		Wan2.1-T2V-14B      &  96.6      & 97.6     & 62.4  &64.9 & 99.2 &98.5 &58.6 & \underline{79.4}    \\
        w/ VideoReward    &   \underline{96.7}            & \textbf{97.9} &62.9    &\underline{66.5} &\underline{99.3} & \underline{98.5} & 41.6 & 78.2        \\
        w/ UnifiedReward-Think     &  96.4            & 97.7 &\underline{63.9}    &65.2 &\textbf{99.4} & 98.4 & \underline{58.3} & 78.4        \\
		\midrule
                
          \textbf{w/ UnifiedReward-Flex}       & \textbf{96.9}            & \underline{97.8}   &\textbf{65.1}   &\textbf{66.9} &\underline{99.3}& \textbf{99.0} & \textbf{70.8} & \textbf{79.9}   \\
          
		\bottomrule

        \toprule
                
		Models  &
		\makecell[c]{\textbf{Color}}
       & \makecell[c]{\textbf{Spatial} \\\textbf{Relationship}}
  &\makecell[c]{\textbf{Scene}} &\makecell[c]{\textbf{Temporal} \\\textbf{Style}}
&\makecell[c]{\textbf{Overall} \\\textbf{Consistency}}
&\makecell[c]{\textbf{Object} \\\textbf{Class}}
&\makecell[c]{\textbf{Multiple} \\\textbf{Objects}}
&\makecell[c]{\textbf{Appearance} \\\textbf{Style}}
       \\
		\midrule

		Wan2.1-T2V-14B     &  87.7     & 72.6     & 28.8  &23.6 & 25.1 &79.1 &61.8 &22.2      \\
        w/ VideoReward       &   \underline{87.8}            & 77.0 & \underline{28.2}    &23.7 &25.3 & \underline{82.1} & \underline{70.2}& 21.0    \\
        w/ UnifiedReward-Think     &   86.1            & \underline{77.3} &27.2    &\underline{23.8} &\underline{25.4} & 78.4& 63.0 & \underline{22.3}       \\
		\midrule
                
            \textbf{w/ UnifiedReward-Flex}     & \textbf{89.6}         & \textbf{80.8}   &\textbf{30.5}   &\textbf{24.2} & \textbf{25.6} & \textbf{83.2} & \textbf{70.6}  & \textbf{22.4}   \\
		\bottomrule

	\end{tabular} \\

\label{tab:vbench_compare}
\end{table*}

%% file: img/video_compare1.tex
\begin{figure}[!th]

    \centering
    \includegraphics[width=1\linewidth]{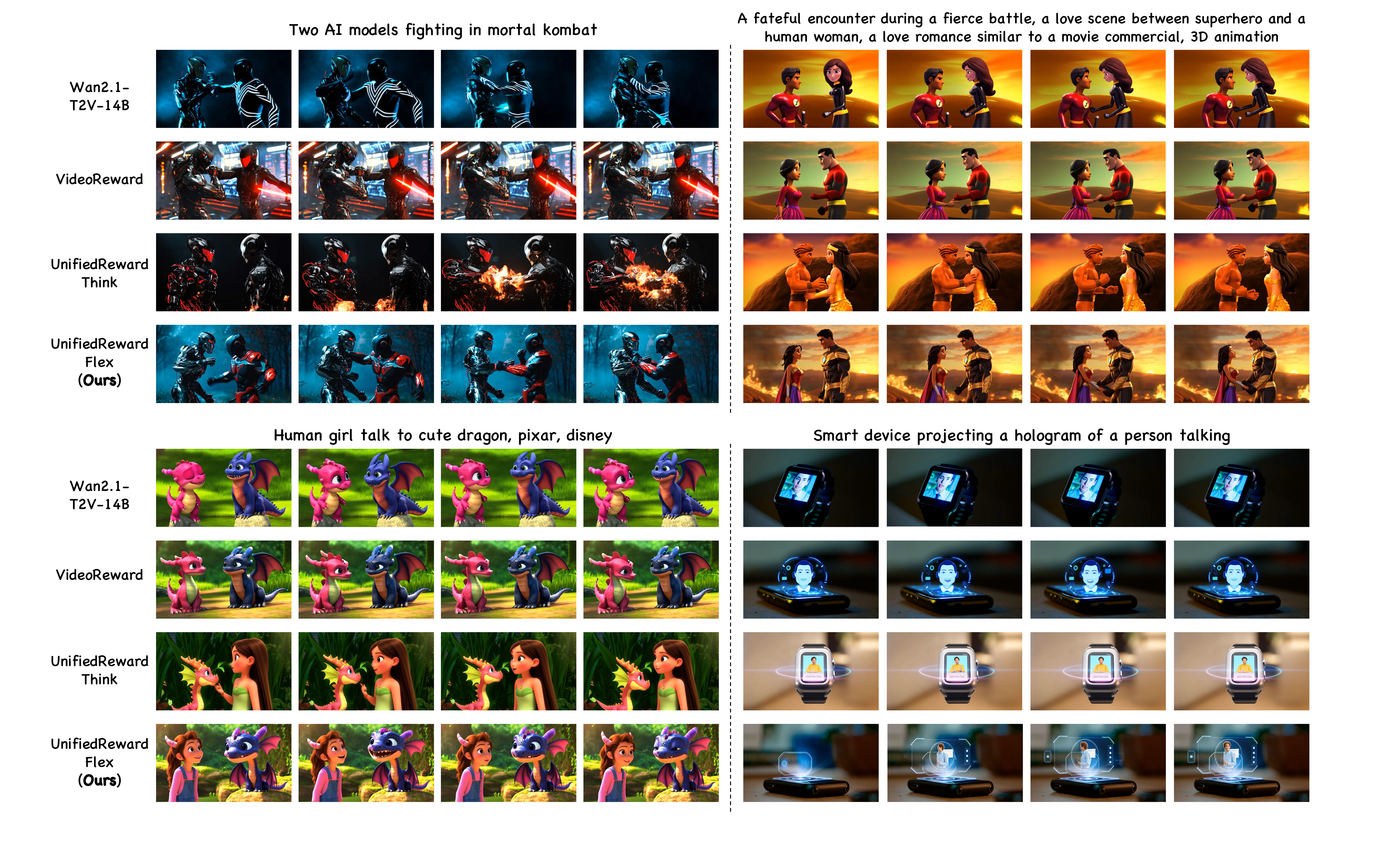}
    \caption{\textbf{Qualitative Comparison on Text-to-Video GRPO (Wan2.1-T2V-14B).}}

    \label{fig:qualitative_video_compare}

\end{figure}

%% file: table/ablation_reward.tex
\begin{table}[t]
\centering
\scriptsize
\setlength{\tabcolsep}{8pt}
\caption{\textbf{Hyperparameter Analysis of $\alpha$}. The best results are in \textbf{bold}, and the second best are \underline{underlined}.}
\begin{tabular}{lccc|lcccccccc}
\toprule
\multirow{2}{*}{\textbf{Model}}& \multicolumn{3}{c}{\textbf{Text-to-Image Generation}} & \multirow{2}{*}{\textbf{Model}} &\multicolumn{3}{c}{\textbf{Text-to-Video Generation}}\\
 \cmidrule(lr){2-4} \cmidrule(lr){6-8}
  & \textbf{UniGenBench}& \textbf{T2I-CompBench} & \multicolumn{1}{c}{\textbf{UnifiedReward}} & & \textbf{Total} & \textbf{Semantic} & \textbf{Quality} \\
\midrule
FLUX.1-dev   & 59.39 &48.57   & 3.07 &  Wan2.1-T2V-14B & 80.81 & 69.66 &  83.60 \\
$\alpha=0$ (w/o $\bar{R}_{\mathrm{dim}}$) & 71.13  & 50.32 &  3.25 & $\alpha=0$ (w/o $\bar{R}_{\mathrm{dim}}$) & 82.46 & 71.79 &  85.13 \\
$\alpha=0.3$ & 72.50 & 50.42 & 3.23 & $\alpha=0.3$ & 82.56 & 72.11 & 85.17\\
$\alpha=0.5$ & 73.10 & 50.90 & \underline{3.29} & $\alpha=0.5$ & 82.82 & 72.34 & 85.44\\
$\alpha=1$ (w/o $R_{\mathrm{overall}}$) & \underline{73.44}  & \textbf{51.59} &   3.26 &$\alpha=1$ (w/o $R_{\mathrm{overall}}$) & \underline{82.89} & \underline{72.42} &  \underline{85.51} \\
\hline
$\alpha=0.7$ (\textbf{Ours}) & \textbf{73.95}&\underline{51.37}   & \textbf{3.31} &  $\alpha=0.7$ (\textbf{Ours}) & \textbf{83.08}  & \textbf{72.94}& \textbf{85.62}\\

\bottomrule
\end{tabular}
\label{tab:efficiency}
\end{table}

%% file: table/flex_scale.tex
\begin{table}[t]
\centering
\small
\renewcommand{\arraystretch}{1.1}
\caption{\textbf{Reward Assessment Performance across Model Scales.}}
\setlength{\tabcolsep}{18.5pt}
\begin{tabular}{c c c c c}
\toprule
\multirow{2}{*}{\textbf{Model}} 
& \multicolumn{2}{c}{\textbf{Image Generation}} 
& \multicolumn{2}{c}{\textbf{Video Generation}}  \\
\cmidrule(lr){2-3}
\cmidrule(lr){4-5}
& \textbf{GenAI-Bench} & \multicolumn{1}{c}{\textbf{MMRB2}} &  \textbf{GenAI-Bench} & \textbf{MJBench} \\
\hline
UnifiedReward-Flex-2B          & 70.3 & 64.6            & 77.5 & 65.2 \\
UnifiedReward-Flex-4B      & 72.1 & 68.5      & 80.2 & 67.8 \\
UnifiedReward-Flex-8B  & \underline{73.4} & \underline{69.2}   
               & \underline{82.5} & \textbf{72.0}\\
UnifiedReward-Flex-32B  & \textbf{74.8} & \textbf{69.9}  & \textbf{82.8} & \underline{71.3} \\

\bottomrule
\end{tabular}
\label{tab:reward_scale}
\end{table}

%% file: img/klein_qualitative_results1.tex
\begin{figure}[!h]

    \centering
    \includegraphics[width=1\linewidth]{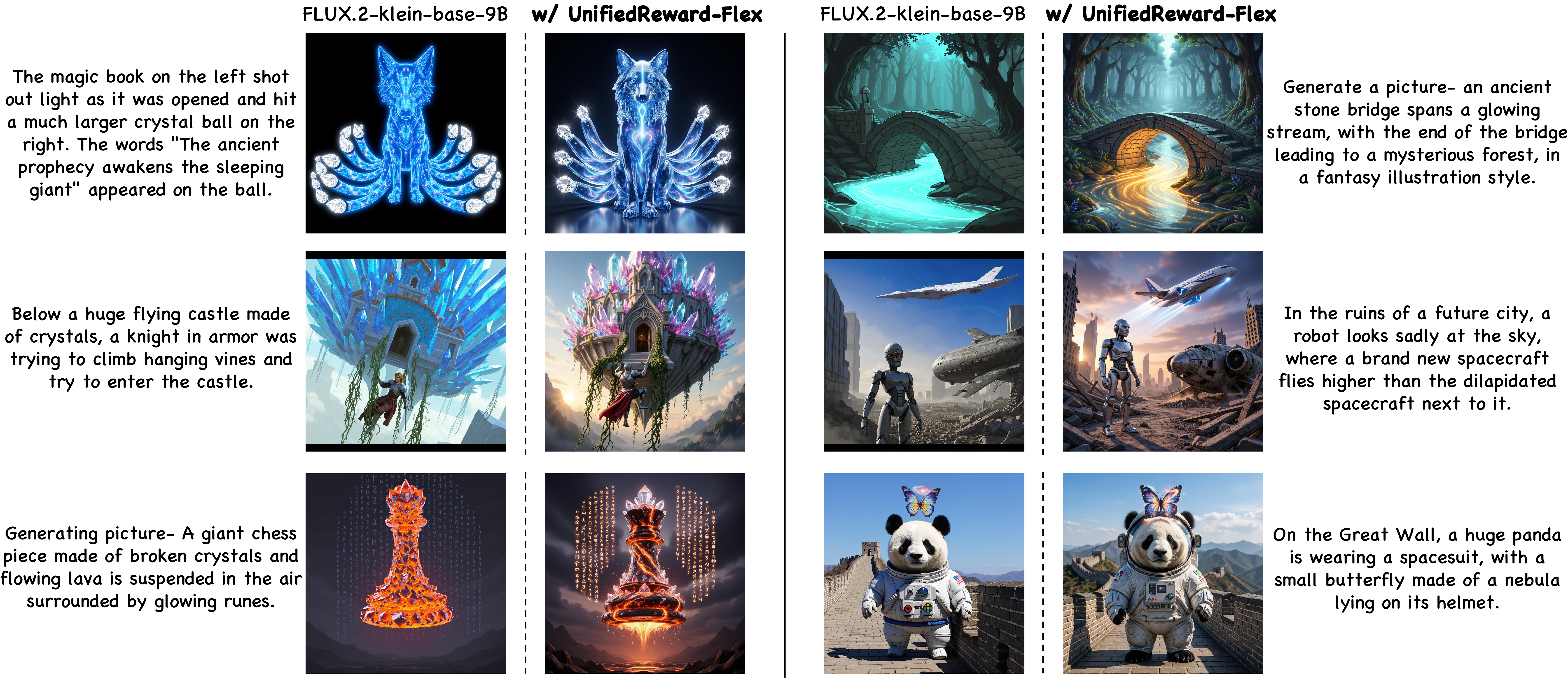}
    \caption{\textbf{Qualitative Results on Text-to-Image GRPO (FLUX.2-klein-base-9B).}
    }

    \label{fig:qualitative_image_compare_klein}

\end{figure}

%% file: table/klein_unigenbench.tex
\begin{table*}[t]
\centering
\scriptsize
\setlength{\tabcolsep}{4pt}
\renewcommand{\arraystretch}{1.2}
\caption{\textbf{Robustness across Different Text-to-Image Generators on UniGenBench.} ``UniGenBench-EvalModel-qwen3vl-32b-v1'' is used as the VLM for evaluation.} 
\begin{tabular}{lc|cccccccccc}
\toprule
\textbf{Model} & \textbf{Overall} & Style & World Know. & Attribute & Action & Relation. & Compound & Grammar & Logic.Reason. & Layout & Text \\
\midrule
FLUX.1-dev        &59.39   & 85.10 & 85.92 & 65.28 & 61.41 & 64.97 & 43.56 & 60.16 & 24.77 & 70.52 & 32.18 \\
\textbf{w/ UnifiedReward-Flex} & \textbf{73.95} &\textbf{90.30}   & \textbf{89.87} & \textbf{79.38} & \textbf{73.38} & \textbf{78.55}& \textbf{69.46} & \textbf{63.10} & \textbf{46.59} & \textbf{79.66} & \textbf{59.20}  \\
\midrule
FLUX.2-klein-base-9B        &78.93   & 97.50 & 91.61 & 83.65 & 77.00 & 86.42 & 78.61 & 76.87 & 53.41 & 88.43 & 55.75 \\
\textbf{w/ UnifiedReward-Flex} & \textbf{81.54} &\textbf{97.60}   & \textbf{91.93} & \textbf{85.47} & \textbf{78.42} & \textbf{86.42}& \textbf{81.96} & \textbf{76.97} & \textbf{58.64} & \textbf{88.43} & \textbf{69.54}  \\
\bottomrule
\label{tab:unigenbench_klein}

\end{tabular} 

\end{table*}

%% file: table/klein_image_eval.tex
\begin{table}[t]
\centering
\scriptsize
\setlength{\tabcolsep}{9pt}
\caption{\textbf{Robustness across Different Text-to-Image Generators on Semantic Consistency and Image Quality.} The best results are in \textbf{bold}, and the second best are \underline{underlined}.}
\begin{tabular}{lccccccccccc}

\toprule
\multirow{2}{*}{\textbf{Model}}& \multicolumn{4}{c}{\textbf{Semantic Consistency}} & \multicolumn{3}{c}{\textbf{Image Quality}}\\
 \cmidrule(lr){2-5} \cmidrule(lr){6-8}
 & \textbf{UniGenBench}& \textbf{T2I-CompBench} &\textbf{GenEval} & \textbf{CLIP} & \textbf{PickScore} & \textbf{UnifiedReward} & \textbf{Aesthetic} \\
\midrule
FLUX.1-dev   & 59.39 &48.57 &  62.18  & 34.40 & 22.70 & 3.07 &  6.13 \\
\textbf{w/ UnifiedReward-Flex} &\textbf{73.95} & \textbf{51.37} &  \textbf{69.62}  & \textbf{36.25} & \textbf{23.42} & \textbf{3.31}& \textbf{6.56}\\
\midrule
FLUX.2-klein-base-9B   & 78.93 &53.72 &  78.99  & 35.59 & 22.51 & 3.81 &  5.89 \\

\textbf{w/ UnifiedReward-Flex} &\textbf{81.54} & \textbf{58.62} &  \textbf{81.22}  & \textbf{36.42} & \textbf{23.07} & \textbf{3.98}& \textbf{6.06}\\

\bottomrule
\label{tab:ood_compare_klein}
\end{tabular}

\end{table}

%% file: table/vbench_wan22.tex
\begin{table*}[!th]
\setlength\tabcolsep{6pt}
\centering
\scriptsize
\caption{\textbf{Robustness across Different Text-to-Video Generators on VBench.} The first seven metrics correspond to the \textit{Quality} type, while the remaining correspond to the \textit{Semantic} type.}

\begin{tabular}{ccccccccccc}

		\toprule
         Models &
	\makecell[c]{\textbf{Subject}\\ \textbf{Consistency}}
      & \makecell[c]{\textbf{Background}\\ \textbf{Consistency}}
  &\makecell[c]{\textbf{Aesthetic} \\\textbf{Quality}} &\makecell[c]{\textbf{Imaging} \\\textbf{Quality}}

&\makecell[c]{\textbf{Temporal} \\\textbf{Flickering}}
&\makecell[c]{\textbf{Motion} \\\textbf{Smoothness}}
& \makecell[c]{\textbf{Dynamic} \\\textbf{Degree}}
& \makecell[c]{\textbf{Human} \\\textbf{Action}}

       \\
		\midrule

		Wan2.1-T2V-14B      &  96.6      & 97.6     & 62.4  &64.9 & 99.2 &98.5 &58.6 & 79.4    \\

        \textbf{w/ UnifiedReward-Flex}       & \textbf{96.9}            & \textbf{97.8}   &\textbf{65.1}   &\textbf{66.9} &\textbf{99.3}& \textbf{99.0} & \textbf{70.8} & \textbf{79.9}   \\

        \midrule
 		Wan2.2-T2V-A14B      & 94.8 & 95.7 & 63.9  &65.8 & 98.8 &97.0 &80.0 & 82.0    \\
                
          \textbf{w/ UnifiedReward-Flex}       & \textbf{94.9}            & \textbf{95.9}   &\textbf{64.7}   &\textbf{66.9} &\textbf{98.8}& \textbf{97.1} & \textbf{80.5} & \textbf{84.6}   \\
          
		\bottomrule

        \toprule
                
		Models  &
		\makecell[c]{\textbf{Color}}
       & \makecell[c]{\textbf{Spatial} \\\textbf{Relationship}}
  &\makecell[c]{\textbf{Scene}} &\makecell[c]{\textbf{Temporal} \\\textbf{Style}}
&\makecell[c]{\textbf{Overall} \\\textbf{Consistency}}
&\makecell[c]{\textbf{Object} \\\textbf{Class}}
&\makecell[c]{\textbf{Multiple} \\\textbf{Objects}}
&\makecell[c]{\textbf{Appearance} \\\textbf{Style}}
       \\
		\midrule

		Wan2.1-T2V-14B     &  87.7     & 72.6     & 28.8  &23.6 & 25.1 &79.1 &61.8 &22.2      \\
        \textbf{w/ UnifiedReward-Flex}     & \textbf{89.6}         & \textbf{80.8}   &\textbf{30.5}   &\textbf{24.2} & \textbf{25.6} & \textbf{83.2} & \textbf{70.6}  & \textbf{22.4}   \\
        \midrule
		Wan2.2-T2V-A14B     &  85.8     & 77.2     & 30.2  &23.4 & 25.1 &80.3 &67.0 &21.4      \\
                
            \textbf{w/ UnifiedReward-Flex}     & \textbf{90.4}         & \textbf{81.1}   &\textbf{32.7}   &\textbf{23.7} & \textbf{25.5} & \textbf{84.9} & \textbf{68.3}  & \textbf{21.7}   \\
		\bottomrule

	\end{tabular} \\

\label{tab:vbench_compare_wan22}
\end{table*}

%% file: img/wan22_qualitative_results1.tex
\begin{figure}[!h]

    \centering
    \includegraphics[width=1\linewidth]{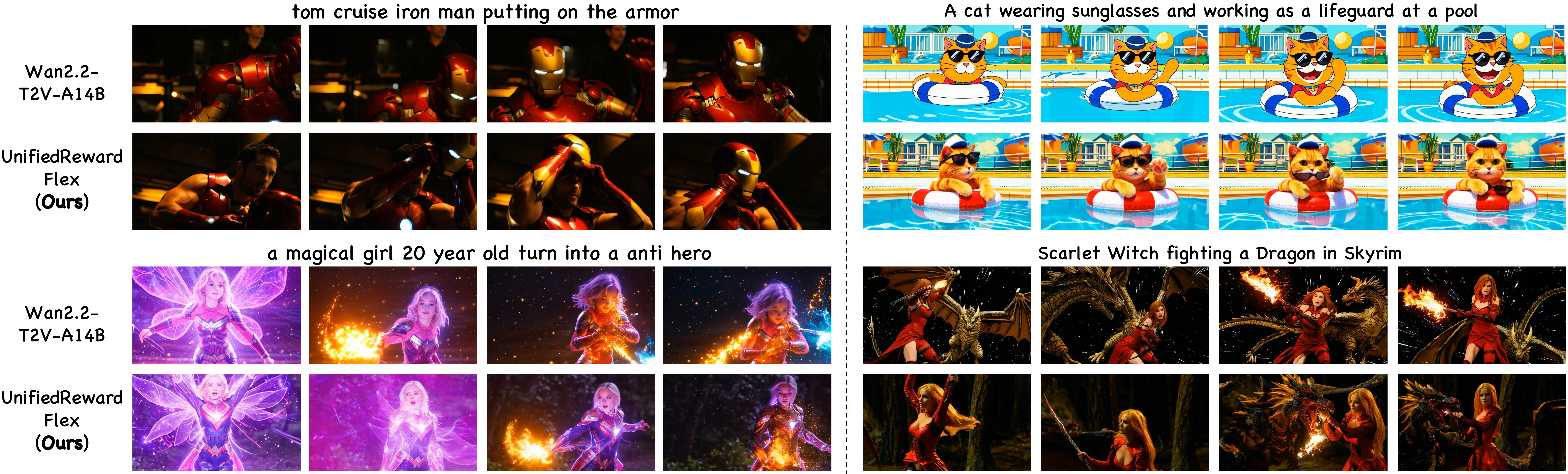}
    \caption{\textbf{Qualitative Results on Text-to-Video GRPO (Wan2.2-T2V-A14B).}
    }

    \label{fig:qualitative_video_compare_wan22}

\end{figure}

%% file: table/efficiency.tex
\begin{table}[t]
\centering
\scriptsize
\setlength{\tabcolsep}{10pt}
\caption{\textbf{Comparison of Training Efficiency (Seconds per Step).}}
\begin{tabular}{lcccccccccccc}
\toprule

  & \textbf{PickScore}& \textbf{HPSv3}& \textbf{UnifiedReward} & \textbf{VideoReward} & \multicolumn{1}{c}{\textbf{UnifiedReward-Think}} & \textbf{UnifiedReward-Flex}  \\
\midrule
FLUX.1-dev  & 102s & 103s & 109s  & --- & 124s & 143s  \\
Wan2.1-T2V-14B &--- & --- & --- & 285s &  328s &  336s  \\

\bottomrule
\end{tabular}
\label{tab:ablation_reward}
\end{table}

%% file: img/image_train_step1.tex
\begin{figure}[!t]

    \centering
    \includegraphics[width=1\linewidth]{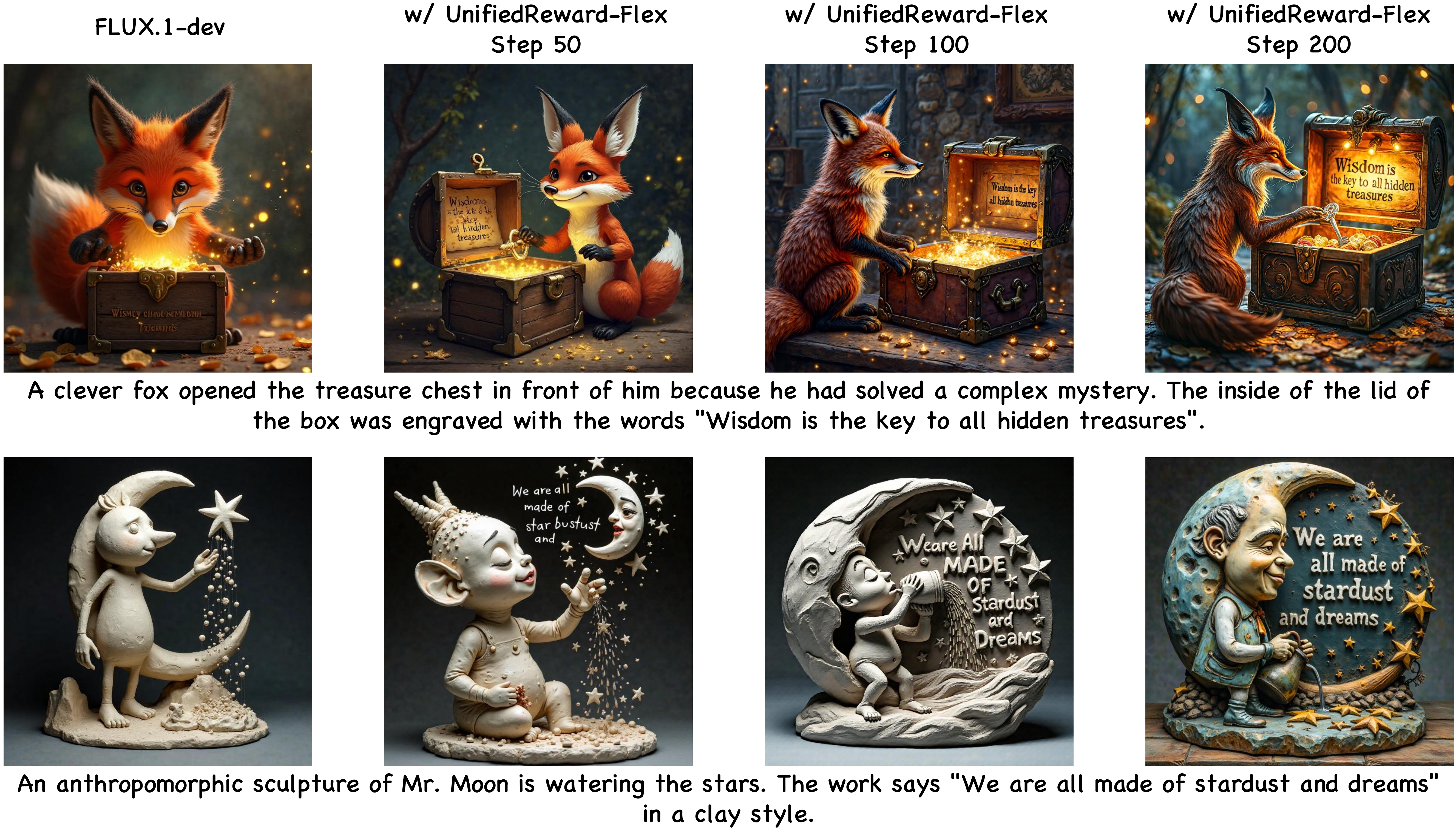}
    \caption{\textbf{Qualitative Results of Text-to-Video Generation during Training Progress.}}

    \label{fig:image_train_step}

\end{figure}

%% file: img/clip_curve1.tex
\begin{figure}[!t]

    \centering
    \includegraphics[width=1\linewidth]{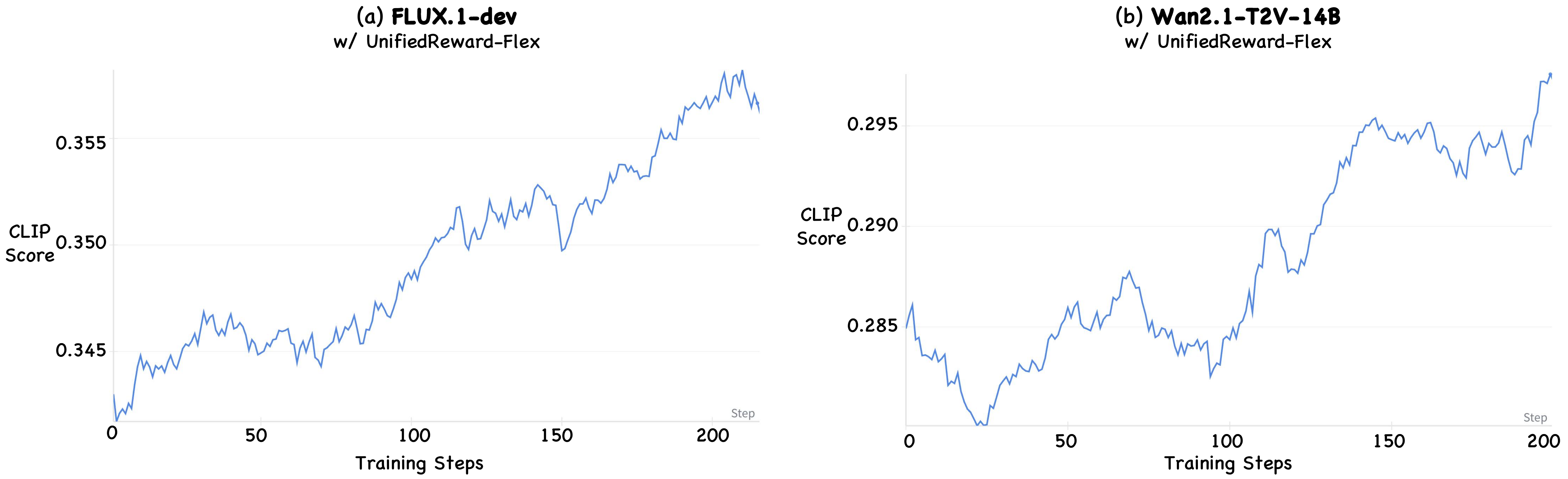}
    \caption{\textbf{Monitoring Training Progress via CLIP Score.}}

    \label{fig:clip_curve}

\end{figure}

%% file: img/video_train_step1.tex
\begin{figure}[!t]

    \centering
    \includegraphics[width=1\linewidth]{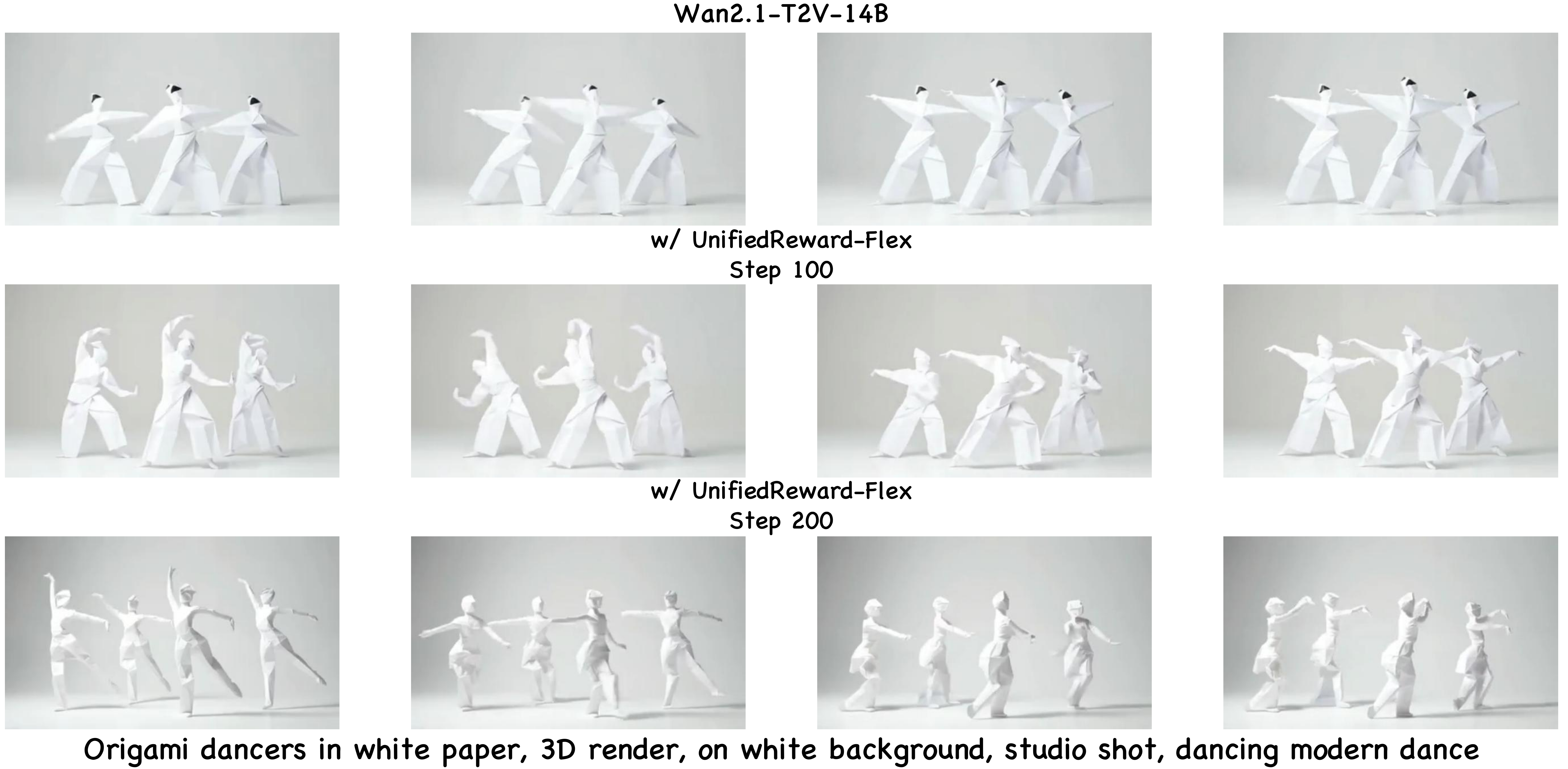}
    \caption{\textbf{Qualitative Results of Text-to-Video Generation during Training Progress.}}

    \label{fig:video_train_step}

\end{figure}

%% file: sections/5conclusion.tex
\section{Conclusion}
This paper introduces \ourmethod, a unified personalized reward model that circumvents the limitations of traditional ``one-size-fits-all'' evaluation in visual generation. By coupling dynamic hierarchical assessment with context-adaptive reasoning, our approach transcends rigid rubrics to capture the nuanced and subjective nature of human preferences. Leveraging a two-stage training pipeline, i.e., structured reasoning distillation and reasoning-aware Direct Preference Optimization (DPO), we demonstrate that \ourmethod yields significantly more precise and context-sensitive reward signals. Empirical validation within the Group Relative Policy Optimization (GRPO) framework reveals that our method consistently outperforms existing baselines, achieving significant improvements in both visual fidelity and semantic alignment for image and video synthesis.

%% file: sections/7appendix.tex
\input{img/supp_image_case_1}
\input{img/supp_video_case_1}

%% file: img/supp_image_case_1.tex
\begin{figure}[!h]

    \centering
    \includegraphics[width=1\linewidth]{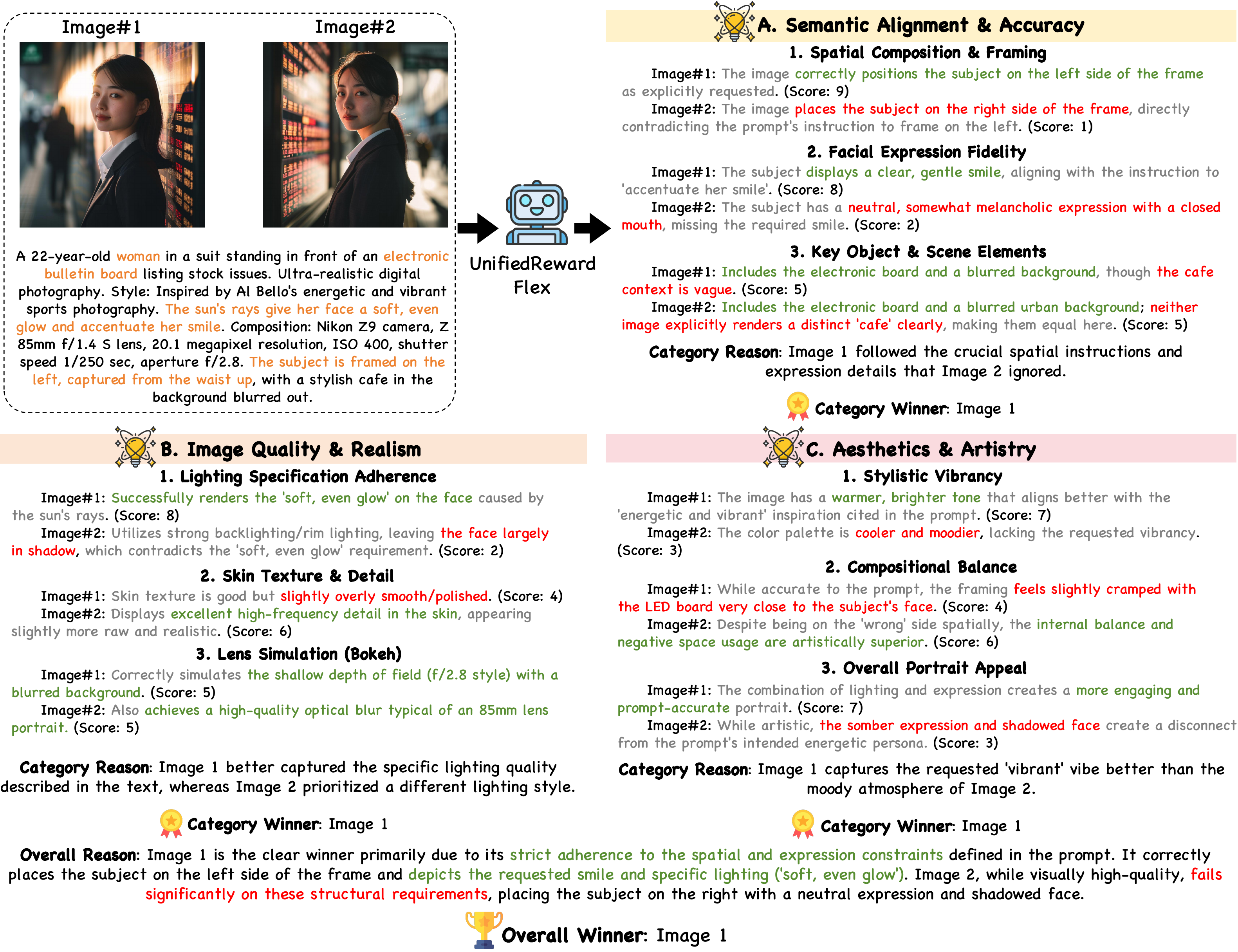}
    \caption{\textbf{More Qualitative Result of \ourmethod on Image Generation Personalized Reward Reasoning.}}

    \label{fig:more_image_case}

\end{figure}

%% file: img/supp_video_case_1.tex
\begin{figure}[!t]

    \centering
    \includegraphics[width=1\linewidth]{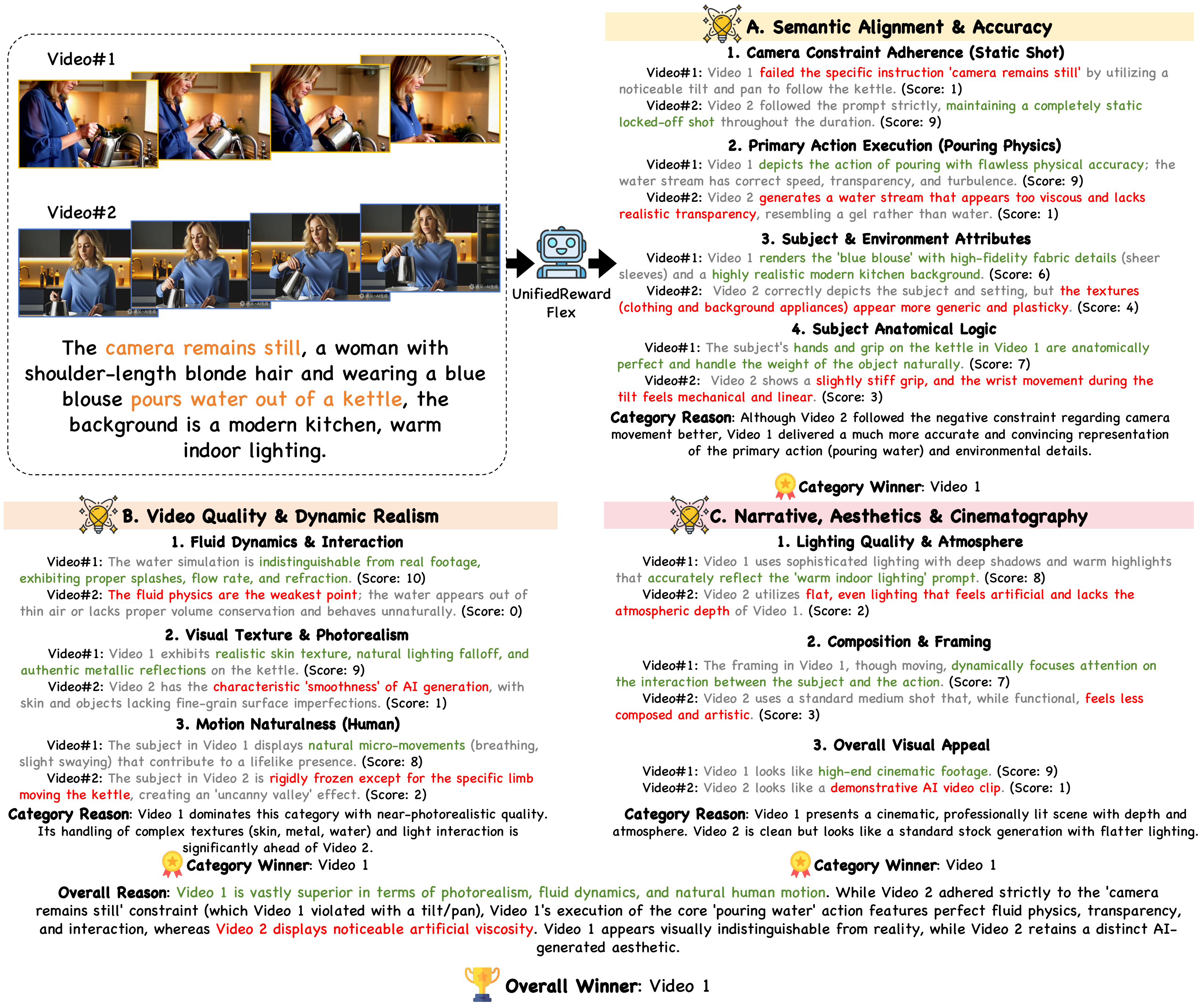}
    \caption{\textbf{More Qualitative Result of \ourmethod on Video Generation Personalized Reward Reasoning.}}

    \label{fig:more_video_case}

\end{figure}